\documentclass[11pt,a4paper]{article}
\usepackage[hyperref]{emnlp2020}
\usepackage{times}
\usepackage{amssymb}
\usepackage{amsmath}
\usepackage{latexsym}
\usepackage[pdftex]{graphicx}
\usepackage{tabularx}

\usepackage{algorithmic}
\usepackage{algorithm}

\definecolor{lightblue}{rgb}{0.19, 0.55, 0.91}
\definecolor{darkyellow}{rgb}{1.0, 0.65, 0.0}
\DeclareMathOperator*{\argmax}{arg\,max}

\usepackage{microtype}

\aclfinalcopy 
\def\aclpaperid{2448} 

\title{Seq2Edits: Sequence Transduction Using Span-level Edit Operations}

\author{Felix Stahlberg and Shankar Kumar \\
Google Research \\
\texttt{\{fstahlberg,shankarkumar\}@google.com}}

\date{}

\begin{document}
\maketitle
\begin{abstract}
We propose Seq2Edits, an open-vocabulary approach to sequence editing for natural language processing (NLP) tasks with a high degree of overlap between input and output texts. In this approach, each sequence-to-sequence transduction is represented as a sequence of edit operations, where each operation either replaces an entire source span with target tokens or keeps it unchanged. We evaluate our method on five NLP tasks (text normalization, sentence fusion, sentence splitting \& rephrasing, text simplification, and grammatical error correction) and report competitive results across the board. For grammatical error correction, our method speeds up inference by up to 5.2x compared to full sequence models because inference time depends on the number of edits rather than the number of target tokens. For text normalization, sentence fusion, and grammatical error correction, our approach improves explainability by associating each edit operation with a human-readable tag. 
\end{abstract}

\section{Introduction}

Neural models that generate a target sequence conditioned on a source sequence were initially proposed for machine translation (MT) \citep{sutskever-nmt,kalchbrenner-blunsom-2013-recurrent-continuous,bahdanau-nmt,transformer}, but are now used widely as a central component of a variety of NLP systems (e.g.\ \citet{seq2seqforsummarization,seq2seqforgec}). \citet{t5} argue that even problems that are traditionally not viewed from a sequence transduction perspective can benefit from massive pre-training when framed as a text-to-text problem. However, for many NLP tasks such as correcting grammatical errors in a sentence, the input and output sequence may overlap significantly. Employing a full sequence model in these cases is often wasteful as most tokens are simply copied over from the input to the output. Another disadvantage of a full sequence model is that it does not provide an explanation for why it proposes a particular target sequence.

In this work, inspired by a recent increased interest in text-editing \citep{dong-etal-2019-editnts,lasertagger,felix,edit-tag}, we propose \emph{Seq2Edits}, a sequence {\em editing} model which is tailored towards problems that require only small changes to the input. Rather than generating the target sentence as a series of tokens, our model predicts a sequence of edit operations that, when applied to the source sentence, yields the target sentence. Each edit operates on a span in the source sentence and either copies, deletes, or replaces it with one or more target tokens. Edits are generated auto-regressively from left to right using a modified Transformer~\citep{transformer} architecture to facilitate learning of long-range dependencies. We apply our edit operation based model to five NLP tasks: text normalization, sentence fusion, sentence splitting \& rephrasing, text simplification, and grammatical error correction (GEC).  Our model is competitive across all of these tasks, and improves the state-of-the-art on text normalization~\citep{rws-rnn-text-norm}, sentence splitting \& rephrasing~\citep{botha-etal-2018-learning}, and the JFLEG test set~\citep{napoles-etal-2017-jfleg} for GEC.

\begin{figure*}[t!]
\centering
\small
\includegraphics[scale=0.35]{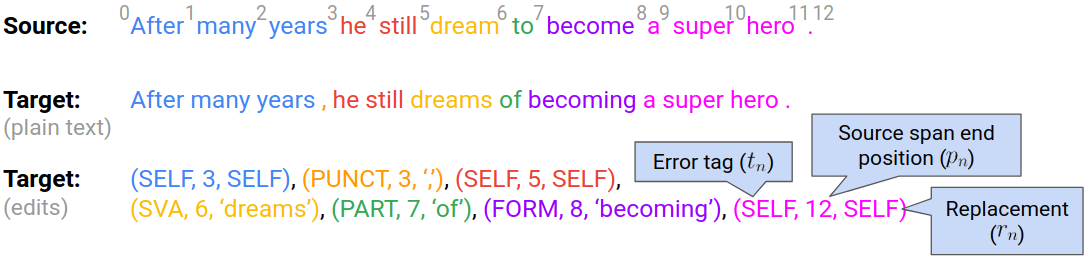}
\caption{Representing grammatical error correction as a sequence of span-based edit operations. The implicit start position for a source span is the end position of the previous edit operation. \texttt{SELF} indicates spans that are copied over from the source sentence ($\mathbf{x}$). The probability of the first two edits is given by:
$P(\text{After many years ,}|\mathbf{x}) =
{\color{lightblue} P(t_1=\text{SELF}|\mathbf{x}) \cdot P(p_1=3|\text{SELF},\mathbf{x}) \cdot P(r_1=\text{SELF}|\text{SELF},3,\mathbf{x})} \cdot
{\color{darkyellow} P(t_2=\text{PUNCT}|\text{SELF},3,\text{SELF},\mathbf{x}) \cdot P(p_2=3|\text{SELF},3,\text{SELF},\text{PUNCT},\mathbf{x})
\cdot P(r_2=\text{,}|\text{SELF},3,\text{SELF},\text{PUNCT},3,\mathbf{x})}
$.
}
\label{fig:edit-ops}
\end{figure*}

Our model is often much faster than a full sequence model for these tasks because its runtime depends on the number of edits rather than the target sentence length. For instance, we report speed-ups of $>$5x on GEC for native English in initial experiments. If applicable, we also predict a task-specific edit-type class (``tag'') along with each edit which explains why that edit was proposed. For example in GEC, the correction of a misspelled word would be labelled with a \texttt{SPELL} (spelling error) whereas changing a word from say first person to third person would be associated with a tag such as \texttt{SVA} (subject-verb agreement error).

\section{Edit-based Sequence Transduction}

\subsection{Representation}
\label{sec:representation}
\label{sec:invalid}

A vanilla sequence-to-sequence (seq2seq) model generates a plain target sequence $\mathbf{y}=y_1^J=y_1, y_2, \dots, y_J\in V^J$ of length $J$ given a source sequence $\mathbf{x}=x_1^I=x_1, x_2, \dots, x_I\in V^I$ of length $I$ over a vocabulary $V$ of tokens (e.g.\ subword units~\citep{sennrich-etal-2016-neural}). For example, in our running grammar correction example in Fig.~\ref{fig:edit-ops}, the source sequence is the ungrammatical sentence $\mathbf{x}=$``{\em After many years he still dream to become a super hero .}'' and the target sequence is the corrected sentence $\mathbf{y}=$``{\em After many years , he still dreams of becoming a super hero .}''. The probability $P(\mathbf{y}|\mathbf{x})$ is factorized using the chain rule:
\begin{equation}
    P(\mathbf{y}|\mathbf{x})=\prod_{j=1}^J P(y_j|y_1^{j-1},\mathbf{x}).
\end{equation}
Instead of predicting the target sequence $\mathbf{y}$ directly, the \emph{Seq2Edits} model predicts a sequence of $N$ edit operations. Each edit operation $(t_n, p_n, r_n)\in T\times \mathbb{N}_0 \times V$ is a 3-tuple that represents the action of replacing the span from positions $p_{n-1}$ to $p_n$ in the source sentence with the replacement token $r_n$ associated with an explainable tag $t_n$ ($T$ is the tag vocabulary).\footnote{In our ablation experiments without tags, each edit operation is represented by a 2-tuple $(p_n,r_n)$ instead.} $t_n=r_n=\mathtt{SELF}$ indicates that the source span is kept as-is. Insertions are modelled with $p_n=p_{n-1}$ that corresponds to an empty source span (see the insertion of ``,'' in Fig.~\ref{fig:edit-ops}), deletions are represented with a special token $r_n=\mathtt{DEL}$. The edit operation sequence for our running example is shown in Fig.~\ref{fig:edit-ops}. The target sequence $\mathbf{y}$ can be obtained from the edit operation sequence using Algorithm~\ref{alg:edit2seq}.

\begin{algorithm}[t!]
\caption{ApplyEdits()}
\label{alg:edit2seq}
\begin{algorithmic}[1]
\STATE{$p_0\gets 0$} \COMMENT{First span starts at 0.}
\STATE{$\mathbf{y}\gets\epsilon$} \COMMENT{Initialize $\mathbf{y}$ with the empty string.}
\FOR{$n\gets 1$ \TO $N$}
  \IF{$t_n = \mathtt{SELF}$}
    \STATE{$\mathbf{y}\gets\text{concat}(\mathbf{y}, x_{p_{n-1}}^{p_n})$}
  \ELSIF {$r_n \neq \mathtt{DEL}$}
    \STATE{$\mathbf{y}\gets\text{concat}(\mathbf{y}, r_n)$}
  \ENDIF
\ENDFOR
\RETURN{$\mathbf{y}$}
\end{algorithmic}
\end{algorithm}

Our motivation behind using span-level edits rather than token-level edits is that the representations are much more compact and easier to learn since local dependencies (within the span) are easier to capture. For some of the tasks it is also more natural to approach the problem on the span-level: a grammatical error is often fixed with more than one (sub)word, and span-level edits retain the language modelling aspect within a span.

Our representation is flexible as it can represent any sequence pair. As an example, a trivial (but not practical) way to construct an edit sequence for any pair $(\mathbf{x},\mathbf{y})$ is to start with a deletion for the entire source sentence $\mathbf{x}$ ($p_1=I$, $r_1=\mathtt{DEL}$) and then insert the tokens in $\mathbf{y}$ ($p_{j+1}=I$, $r_{j+1}=y_j$ for $j\in[1,J]$).

Edit sequences are valid iff.\ spans are in a monotonic left-to-right order and the final span ends at the end of the source sequence, i.e.:
\begin{equation}
p_N=I \land \forall n\in[1,N): p_n \leq p_{n+1}
\end{equation}
None of our models produced invalid sequences at inference time even though we did not implement any feasibility constraints.

\subsection{Inference}
\label{sec:inference}

The output of the edit operation model is a sequence of 3-tuples rather than a sequence of tokens. The probability of the output is computed as:
\begin{equation}
\begin{split}
    P(\mathbf{y}|\mathbf{x}) &= P(t_1^N,p_1^N,r_1^N|\mathbf{x}) \\
    &= \prod_{n=1}^N P(t_n,p_n,r_n|t_1^{n-1},p_1^{n-1},r_1^{n-1},\mathbf{x}).
\end{split}
\end{equation}
For inference, we factorize the conditional probabilities further as:
\begin{equation}
\begin{split}
    P&(t_n,p_n,r_n|t_1^{n-1},p_1^{n-1},r_1^{n-1},\mathbf{x}) \\ 
    =& P(t_n|t_1^{n-1},p_1^{n-1},r_1^{n-1},\mathbf{x}) \\
    & \cdot P(p_n|t_1^{n},p_1^{n-1},r_1^{n-1},\mathbf{x}) \\
    & \cdot P(r_n|t_1^{n},p_1^{n},r_1^{n-1},\mathbf{x}).
\end{split}
\end{equation}
The decoding problem can thus be written as a flat product of conditional probabilities that correspond to tag, span and replacement predictions, that are interleaved:
\begin{equation}
\begin{split}
    \argmax_{N, t_1^N, p_1^N, r_1^N} P(t_1|\mathbf{x})\cdot P(p_1|t_1,\mathbf{x})\cdot P(r_1|t_1,p_1,\mathbf{x})\\
    \cdot P(t_2|t_1,p_1,r_1,\mathbf{x}) \cdots P(r_N|t_1^N,p_1^N,r_1^{N-1},\mathbf{x}).
\end{split}
\end{equation}
At inference time we perform beam decoding over this flat factorization to search for the most likely edit operation sequence. In practice, we scale the scores for the different target features:
\begin{equation}
\begin{split}
\argmax_{N, t_1^N, p_1^N, r_1^N} \sum_{n=1}^N & \lambda_t\log P(t_n|t_1^{n-1},p_1^{n-1},r_1^{n-1},\mathbf{x}) \\
    & + \lambda_p\log P(p_n|t_1^{n},p_1^{n-1},r_1^{n-1},\mathbf{x}) \\
    & + \lambda_r\log P(r_n|t_1^{n},p_1^{n},r_1^{n-1},\mathbf{x}).
\end{split}
\end{equation}
where the three scaling factors $\lambda_t$, $\lambda_p$, $\lambda_r$ are optimized on the respective development set.

\subsection{Neural Architecture}

\begin{figure}[t!]
\centering
\small
\includegraphics[scale=0.26]{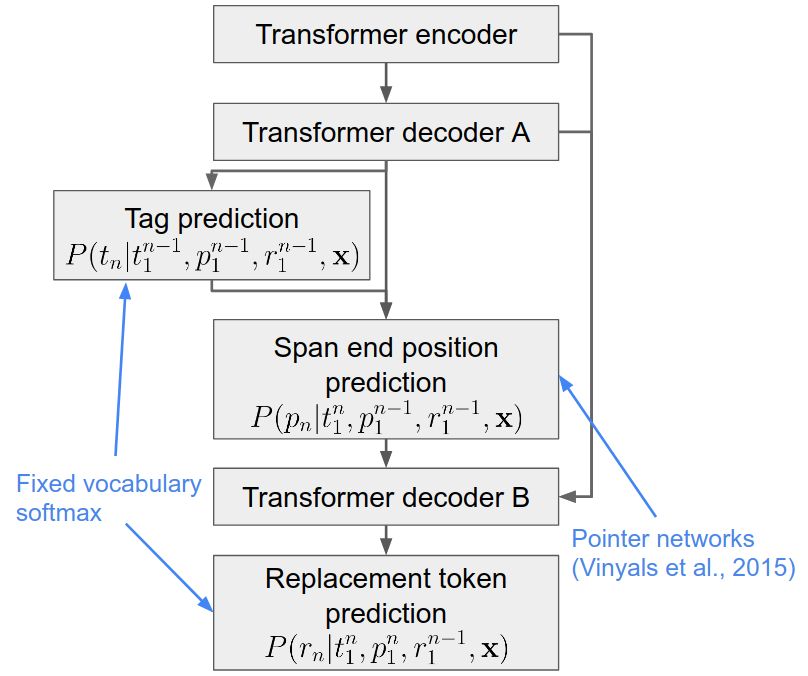}
\caption{\emph{Seq2Edits} consists of a Transformer~\citep{transformer} encoder and a Transformer decoder that is divided horizontally into two parts (A and B). The tag and span predictions are located in the middle of the decoder layer stack between both parts. A single step of prediction is shown.}
\label{fig:architecture}
\end{figure}

Our neural model (illustrated in Fig.~\ref{fig:architecture}) is a generalization of the original Transformer architecture of \citet{transformer}. Similarly to the standard Transformer we feed back the predictions of the previous time step into the Transformer decoder (A). The feedback loop at time step $n$ is implemented as the concatenation of an embedding of $t_{n-1}$, the $p_{n-1}$-th encoder state, and an embedding of $r_{n-1}$. The Transformer decoder A is followed by a cascade of tag prediction and span end position prediction. We follow the idea of pointer networks~\citep{pointer-networks} to predict the source span end position using the attention weights over encoder states as probabilities. The input to the pointer network are the encoder states (keys and values) and the output of the previous decoder layer (queries). The span end position prediction module is a Transformer-style \citep{transformer} single-head attention (``scaled dot-product'') layer over the encoder states:
\begin{equation}
   P(p_n|t_1^{n},p_1^{n-1},r_1^{n-1},\mathbf{x}) =  \text{softmax}(\frac{QK^T}{\sqrt{d}}),
\label{eq:pointer}
\end{equation}
where $Q$ is a $d$-dimensional linear transform of the previous decoder layer output at time step $n$, $K\in\mathbb{R}^{I\times d}$ is a linear transform of the encoder states, and $d$ is the number of hidden units.

\begin{table}
\centering
\small
\begin{tabular}{lll}
\hline \textbf{Hyper-parameter} &  \textbf{Base} &  \textbf{Big} \\ \hline
Hidden units & 512 & 1,024 \\
Encoder layers & 6 & 6 \\
Decoder A layers & 3 & 4 \\
Decoder B layers & 3 & 4 \\
\hline
No.\ of parameters (w/o embeddings) & 53M & 246M \\
\end{tabular}
\caption{\label{tab:base-big} The ``Base'' and ``Big'' configurations.}
\end{table}

\begin{table*}
\centering
\small
\begin{tabular}{@{\hspace{0em}}l@{\hspace{0em}}c@{\hspace{1em}}ccccc@{\hspace{0em}}}
\hline  &  \multicolumn{2}{c}{\textbf{Text normalization}} & \textbf{Sentence} & \textbf{Sentence} & \textbf{Simplification} & \textbf{Grammar} \\ 
&  \textbf{English} & \textbf{Russian} & \textbf{fusion} & \textbf{splitting} & & \textbf{correction} \\ \hline
Training data & \multicolumn{2}{c}{Wikipedia} & DiscoFuse & WikiSplit & WikiLarge & Lang-8, FCE, \\
 &  & &  &  &  & W\&I \\
Number of sentences & 881K & 816K & 4.5M & 990K & 296K & 2M \\
Task-specific tags & \multicolumn{2}{c}{Semiotic class} & Type & - & - & Error tag \\
Task-specific tag vocabulary size & 18 & 15 & 15 & - & - & 28 \\
Source tokenization & \multicolumn{2}{c}{Characters} & Subwords & Subwords & Subwords & Subwords \\
Target tokenization & \multicolumn{2}{c}{Words} & Subwords & Subwords & Subwords & Subwords \\
Fraction of changed tokens per sentence & 9.9\% & 15.7\% & 13.9\% & 8.7\% & 52.0\% & 10.5\% \\
Average source length in tokens ($I$) & 64.9 & 82.4 & 36.9 & 39.8 & 31.7 & 14.4 \\
Average target length in tokens ($J$) & 57.3 & 69.4 & 31.8 & 36.5 & 15.8 & 13.0 \\
Average number of span-level edits ($N$) & 9.6 & 14.2 & 7.4 & 11.8 & 13.1 & 4.7 \\
\hline
\end{tabular}
\caption{\label{tab:data} Statistics for the task-specific training data. The $I$, $J$, and $N$ variables are introduced in Sec.~\ref{sec:representation}. Our subword-based systems use the implementation available in Tensor2Tensor~\citep{vaswani-etal-2018-tensor2tensor} with a vocabulary size of 32K. The pre-training data is described in the text. See Appendix \ref{sec:tags} for the full tag vocabularies.}
\end{table*}

A 6-dimensional embedding of the predicted tag $t_n$ and the encoder state corresponding to the source span end position $p_n$ are fed into yet another Transformer decoder (B) that predicts the replacement token $r_n$. Alternatively, one can view A and B as a single Transformer decoder layer stack, in which we added the tag prediction and the span end position prediction as additional layers in the middle of that stack. We connect the decoders A and B with residual connections~\citep{residual-connections} to facilitate learning. The network is trained by optimizing the sum of three cross-entropies that correspond to tag prediction, span prediction, and replacement token prediction, respectively.\footnote{We use an unweighted sum as we did not observe gains from weighting the three losses during training.} In our experiments without tags we leave out the tag prediction layer and the loss computed from it. We experiment with two different model sizes: ``Base'' and ``Big''. The hyper-parameters for both configurations are summarized in Table~\ref{tab:base-big}. Hyper-parameters not listed here are borrowed from the {\it transformer\_clean\_base} and {\it transformer\_clean\_big} configurations in the Tensor2Tensor toolkit~\citep{vaswani-etal-2018-tensor2tensor}.

\begin{table*}
\centering
\small
\begin{tabular}{lcccc}
\hline \textbf{Task} & \textbf{Feature weights ($\lambda$)} & \textbf{Iterative refinement} & \textbf{Length normalization} & \textbf{Identity penalty} \\ \hline
Text normalization & $\checkmark$ &  & & \\
Sentence fusion & $\checkmark$ & & & \\
Sentence splitting & $\checkmark$ & & & \\
Simplification & $\checkmark$ & $\checkmark$ & $\checkmark$ & $\checkmark$ \\
Grammar correction & $\checkmark$ & $\checkmark$ &  & $\checkmark$ \\
\hline
\end{tabular}
\caption{\label{tab:tunable-params} Decoding parameters that are tuned on the respective development sets. Feature weight tuning refers to the $\lambda$-parameters in Sec.~\ref{sec:inference}. We use the length normalization scheme from \citet{gnmt} with the parameter $\alpha$.}
\end{table*}

\section{Experiments}
\label{sec:experiments}

We evaluate our edit model on five NLP tasks:\footnote{Citations in this bullet list refer to the test sets we used and do not necessarily point to the pioneering works.}

\begin{itemize}
    \item Text normalization for speech applications~\citep{rws-rnn-text-norm} -- converting number expressions such as ``{\em 123}'' to their verbalizations (e.g. ``{\em one two three}'' or ``{\em one hundred twenty three}'', etc.) depending on the context.
    \item Sentence fusion~\citep{geva-etal-2019-discofuse} -- merging two independent sentences to a single coherent one, e.g.\ ``{\em I need his spirit to be free. I can leave my body.}'' $\rightarrow$ ``{\em I need his spirit to be free so that I can leave my body.}''
    \item Sentence splitting \& rephrasing~\citep{botha-etal-2018-learning} -- splitting a long sentence into two fluent sentences, e.g.\ ``{\em Bo Saris was born in Venlo, Netherlands, and now resides in London, England.}'' $\rightarrow$ ``{\em Bo Saris was born in Venlo , Netherlands. He currently resides in London, England.}''
    \item Text simplification~\citep{zhang-lapata-2017-sentence} -- reducing the linguistic complexity of text, e.g.\ ``{\em The family name is derived from the genus Vitis.}'' $\rightarrow$ ``{\em The family name comes from the genus Vitis.}''
    \item Grammatical error correction~\citep{ng-etal-2014-conll,napoles-etal-2017-jfleg,bryant-etal-2019-bea} -- correcting grammatical errors in written text, e.g.\ ``{\em In a such situaction}'' $\rightarrow$ ``{\em In such a situation}''.
\end{itemize}

Our models are trained on packed examples (maximum length: 256) with Adafactor~\citep{adafactor} using the Tensor2Tensor~\citep{vaswani-etal-2018-tensor2tensor} library. We report results both with and without pre-training. Our pre-trained models for all tasks are trained for 1M iterations on 170M sentences extracted from English Wikipedia revisions and 176M sentences from English Wikipedia round-trip translated via German~\citep{lichtarge-etal-2019-corpora}. For grammatical error correction we use ERRANT~\citep{bryant-etal-2017-automatic,felice-etal-2016-automatic} to derive span-level edits from the parallel text. On other tasks we use a minimum edit distance heuristic to find a token-level edit sequence and convert it to span-level edits by merging neighboring edits.

\begin{table*}
\centering
\small
\begin{tabular}{@{\hspace{0em}}l@{\hspace{0em}}c@{\hspace{1em}}c@{\hspace{1em}}c@{\hspace{1em}}c@{\hspace{1em}}cccccccc}
\hline &  \textbf{Model} & \textbf{Tags} & \textbf{Tuning} & \textbf{Pre-} & \multicolumn{2}{c}{\textbf{Text norm.\ (SER$\downarrow$)}} & \textbf{Fusion} & \textbf{Splitting} & \textbf{Simpl.} & \multicolumn{3}{c}{\textbf{Grammar (BEA-dev)}} \\ 
& \textbf{size} &  & & \textbf{training} & \textbf{English} &  \textbf{Russian} & \textbf{(SARI$\uparrow$)} & \textbf{(SARI$\uparrow$)} & \textbf{(SARI$\uparrow$)} & \textbf{P$\uparrow$} & \textbf{R$\uparrow$} & \textbf{F$_{0.5}\uparrow$} \\ \hline
\footnotesize{a} & Base &  & & & 1.40 & 4.13 & 87.15 & 58.9 & 31.87 & 23.3 & 11.0 & 19.0 \\
\footnotesize{b} & Base & $\checkmark$ & & & 1.36 & 3.95 & 87.33 & 58.9 & 32.10 & 22.5 & 13.3 & 19.8 \\
\hline
\footnotesize{c} & Big &  & & $\checkmark$ & - & - & 88.77 & 63.5 & 33.01 & 50.1 & 34.4 & 45.9 \\
\footnotesize{d} & Big & $\checkmark$ & & $\checkmark$ & - & - & 88.67 & 63.6 & 34.54 & 53.7 & 35.3 & 48.6 \\
\hline
\footnotesize{e} & Big &  & $\checkmark$ & $\checkmark$ & - & - & 88.73 & 63.6 & 37.16 & 49.0 & 38.6 & 46.5 \\
\footnotesize{f} & Big & $\checkmark$ & $\checkmark$ & $\checkmark$ & - & - & 88.72 & 63.6 & 36.30 & 50.9 & 39.1 & 48.0 \\
\hline
\end{tabular}
\caption{\label{tab:single-model-all-tasks} Single model results. For metrics marked with ``$\uparrow$'' (SARI, P(recision), R(ecall), F$_{0.5}$) high scores are favorable, whereas the sentence error rate (SER) is marked with ``$\downarrow$'' to indicate the preference for low values. Tuning refers to optimizing the decoding parameters listed in Table~\ref{tab:tunable-params} on the development sets.}
\end{table*}

The task-specific data is described in Table~\ref{tab:data}. The number of iterations in task-specific training is set empirically based on the performance on the development set (between 20K-75K for fine-tuning, between 100K-300K for training from scratch). Fine-tuning is performed with a reduced learning rate of $3 \times 10^{-5}$.  For each task we tune a different set of decoding parameters  (Table~\ref{tab:tunable-params}) such as the $\lambda$-parameters from Sec.~\ref{sec:inference}, on the respective development sets. For text simplification and grammatical error correction, we perform multiple beam search passes (between two and four) by feeding back the output of beam search as input to the next beam search pass. This is very similar to the iterative decoding strategies used by~\citet{lichtarge-etal-2019-corpora,edit-tag,ms-empirical-human,grundkiewicz-etal-2019-neural} with the difference that we pass through $n$-best lists between beam search passes rather than only the single best hypothesis. During iterative refinement we follow \citet{lichtarge-etal-2019-corpora} and multiply the score of the identity mapping by a tunable identity penalty to better control the trade-off between precision and recall. We use a beam size of 12 in our rescoring experiments in Table~\ref{tab:gec-compare-ensemble} and a beam size of 4 otherwise.

Table~\ref{tab:single-model-all-tasks} gives an overview of our results on all tasks. The tag set consists of semiotic class labels~\citep{rws-rnn-text-norm} for text normalization, discourse type~\citep{geva-etal-2019-discofuse} for sentence fusion, and ERRANT~\citep{bryant-etal-2017-automatic,felice-etal-2016-automatic} error tags for grammatical error correction.\footnote{Appendix \ref{sec:tags} lists the full task-specific tag vocabularies.} For the other tasks we use a trivial tag set: \texttt{SELF}, \texttt{NON\_SELF}, and \texttt{EOS} (end of sequence). We report sentence error rates (SERs$\downarrow$) for text normalization, SARI$\uparrow$ scores~\citep{xu-etal-2016-optimizing} for sentence fusion, splitting, and simplification, and ERRANT~\citep{bryant-etal-2017-automatic} span-based P(recision)$\uparrow$, R(ecall)$\uparrow$, and F$_{0.5}$-scores on the BEA development set for grammar correction.\footnote{Metrics are marked with ``$\uparrow$'' if high values are preferred and with ``$\downarrow$'' if low values are preferred.}

Text normalization is not amenable to our form of pre-training as it does not use subword units and it aims to generate vocalizations rather than text like in our pre-training data. All other tasks, however, benefit greatly from pre-training (compare rows {\bf a} \& {\bf b} with rows {\bf c} \& {\bf d} in Table~\ref{tab:single-model-all-tasks}). Pre-training yields large gains for grammar correction as the pre-training data was specifically collected for this task~\citep{lichtarge-etal-2019-corpora}. Tuning the decoding parameters (listed in Table~\ref{tab:tunable-params}) gives improvements for tasks like simplification, but is less crucial on sentence fusion or splitting (compare rows {\bf c} \& {\bf d} with rows {\bf e} \& {\bf f} in Table~\ref{tab:single-model-all-tasks}). Using tags is especially effective if non-trivial tags are available (compare rows {\bf a} with {\bf b}, {\bf c} with {\bf d}, and {\bf e} with {\bf f} for text normalization and grammar correction).

We next situate our best results (big models with pre-training in rows {\bf e} and {\bf f} of Table~\ref{tab:single-model-all-tasks}) in the context of related work.

\subsection{Text Normalization}

\begin{table}
\centering
\small
\begin{tabular}{lcc}
\hline \textbf{System} & \multicolumn{2}{c}{\textbf{SER$\downarrow$}} \\
& \textbf{English} & \textbf{Russian}  \\ \hline
\citet{mansfield-etal-2019-neural}$^*$ & 2.77 & - \\
\citet{zhang-cl-textnorm} & 1.80 & 4.20 \\
\hline
\textbf{This work (semiotic tags)} & \textbf{1.36} & \textbf{3.95} \\
\hline
\end{tabular}
\caption{\label{tab:textnorm-compare} Sentence error rates on the English and Russian text normalization test sets of \citet{rws-rnn-text-norm}. $^*$: best system from \citet{mansfield-etal-2019-neural} without access to gold semiotic class labels.}
\end{table}

Natural sounding speech synthesis requires the correct pronunciation of numbers based on context e.g. whether the string {\em 123} should be spoken as {\em one hundred twenty three}
or {\em one two three}. Text normalization converts the textual representation
of numbers or other semiotic classes such as abbreviations to their spoken form while both
conveying meaning and morphology~\citep{zhang-cl-textnorm}. The problem is highly context-dependent as abbreviations and numbers often have different feasible vocalizations. Context-dependence is even more pronounced in languages like Russian in which the number words need to be inflected to preserve agreement with context words.

We trained our models on the English and Russian data provided by \citet{rws-rnn-text-norm}. Similarly to others~\citep{zhang-cl-textnorm,rws-rnn-text-norm} we use characters on the input but full words on the output side. Table~\ref{tab:textnorm-compare} shows that our models perform favourably when compared to other systems from the literature on both English and Russian. Note that most existing neural text normalization models \citep{zhang-cl-textnorm,rws-rnn-text-norm} require pre-tokenized input\footnote{Each token in this context is a full semiotic class instance, for example a full date or money expression.} whereas our edit model operates on the untokenized input character sequence. This makes our method attractive for low resource languages where high-quality tokenizers may not be available.

\begin{table}
\centering
\small
\begin{tabular}{lcc}
\hline \textbf{System} & \textbf{Exact$\uparrow$} & \textbf{SARI$\uparrow$}  \\ \hline
\citet{lasertagger} & 53.80 & 85.45 \\
\citet{felix} & 61.31 & 88.78 \\
\citet{bert2bert} & \textbf{63.90} & \textbf{89.52} \\
\hline
\textbf{This work (no tags)} & 61.71 & 88.73 \\
\hline
\end{tabular}
\caption{\label{tab:fusion-compare} Sentence fusion results on the DiscoFuse \citep{geva-etal-2019-discofuse} test set.}
\end{table}

\subsection{Sentence Fusion}

Sentence fusion is the task of merging two independent sentences into a single coherent text and has applications in several NLP areas such as dialogue systems and question answering~\citep{geva-etal-2019-discofuse}.  Our model is on par with the \textsc{Felix} tagger~\citep{felix} on the DiscoFuse dataset~\citep{geva-etal-2019-discofuse} but worse than the \textsc{Bert2Bert} model of \citet{bert2bert} (Table~\ref{tab:fusion-compare}). We hypothesize that \textsc{Bert2Bert}'s strategy of making use of target-side pre-training under a language model objective via BERT~\citep{devlin-etal-2019-bert} is particularly useful for sentence fusion.

\subsection{Sentence Splitting \& Rephrasing}

\begin{table}
\centering
\small
\begin{tabular}{lcc}
\hline \textbf{System} & \textbf{Exact$\uparrow$} & \textbf{SARI$\uparrow$}  \\ \hline
\citet{botha-etal-2018-learning} & 14.6 & 60.6 \\
\citet{lasertagger} & 15.2 & 61.7 \\
\citet{lasertagger} - $\textsc{seq2seq}_\textsc{bert}$ & 15.1 & 62.3 \\
\hline
\textbf{This work (no tags)} & \textbf{17.0} & \textbf{63.6} \\
\hline
\end{tabular}
\caption{\label{tab:split-compare} Sentence splitting results~\citep{botha-etal-2018-learning}.}
\end{table}

Sentence splitting is the inverse task of sentence fusion: Split a long sentence into two fluent shorter sentences. Our models are trained on the WikiSplit dataset~\citep{botha-etal-2018-learning} extracted from the edit history of Wikipedia articles. In addition to SARI scores we report the number of exact matches in Table~\ref{tab:split-compare}. Our edit-based model achieves a higher number of exact matches and a higher SARI score compared to prior work on sentence splitting.

\begin{table}[t!]
\centering
\small
\begin{tabular}{lc}
\hline \textbf{System} &  \textbf{SARI$\uparrow$}  \\ \hline
\citet{lasertagger}  & 32.31 \\
\citet{dong-etal-2019-editnts} & 34.94 \\
\citet{xu-etal-2016-optimizing} & 37.94 \\
\citet{felix}  & \textbf{38.13} \\
\hline
\textbf{This work (no tags)} & 37.16 \\
\hline
\end{tabular}
\caption{\label{tab:simplification-compare} Text simplification results.\footnotemark}
\end{table}

\subsection{Text Simplification}

Our text simplification training set (the WikiLarge corpus \citep{zhang-lapata-2017-sentence}) consists of 296K examples, the smallest amongst all our training corpora. Table~\ref{tab:simplification-compare} shows that our model is competitive, demonstrating that it can benefit from even limited quantities of training data. However, it does not improve the state of the art on this test set.

\footnotetext{We report SARI scores as (re)computed by \citet{felix} for all systems in Table~\ref{tab:simplification-compare} to ensure comparability.}

\begin{table*}
\centering
\small
\begin{tabular}{lccccccc}
\hline \textbf{System} & \multicolumn{3}{c}{\textbf{BEA-dev}}  & \multicolumn{3}{c}{\textbf{CoNLL-14}} & \textbf{JFLEG}  \\
& \textbf{P$\uparrow$} & \textbf{R$\uparrow$} & \textbf{F$_{0.5}\uparrow$} & \textbf{P$\uparrow$} & \textbf{R$\uparrow$} & \textbf{F$_{0.5}\uparrow$} & \textbf{GLEU$\uparrow$}  \\ \hline
\citet{lichtarge-etal-2019-corpora} & - & - & - & 65.5 & 37.1 & 56.8 & 61.6 \\
\citet{edit-tag} & - & - & - & 66.1 & 43.0 & 59.7 & 60.3 \\
\citet{zhao-etal-2019-improving} & - & - & - & 67.7 & 40.6 & 59.8 & - \\
\citet{choe-etal-2019-neural} & 54.4 & 32.2 & 47.8 & - & - & - & - \\
\citet{grundkiewicz-etal-2019-neural} & \textbf{56}.1 & 34.8 & \textbf{50.0} & - & - & - & - \\
\citet{gec-pseudo-data} & - & - & - & \textbf{67.9} & 44.1 & \textbf{61.3} & 59.7 \\
\hline
\textbf{This work (ERRANT tags)} & 50.9 & \textbf{39.1} & 48.0 & 63.0 & \textbf{45.6} & 58.6 & \textbf{62.7} \\
\hline
\end{tabular}
\caption{\label{tab:gec-compare-single-model} Single model results for grammatical error correction.}
\end{table*}

\begin{table*}
\centering
\small
\begin{tabular}{lccccccc}
\hline \textbf{System} & \multicolumn{3}{c}{\textbf{BEA-test}}  & \multicolumn{3}{c}{\textbf{CoNLL-14}} & \textbf{JFLEG}  \\
& \textbf{P$\uparrow$} & \textbf{R$\uparrow$} & \textbf{F$_{0.5}\uparrow$} & \textbf{P$\uparrow$} & \textbf{R$\uparrow$} & \textbf{F$_{0.5}\uparrow$} & \textbf{GLEU$\uparrow$}  \\ \hline
\citet{lichtarge-etal-2019-corpora} & - & - & - & 66.7 & 43.9 & 60.4 & 63.3 \\
\citet{edit-tag} & - & - & - & 68.3 & 43.2 & 61.2 & 61.0 \\
\citet{zhao-etal-2019-improving} & - & - & - & 71.6 & 38.7 & 61.2 & 61.0 \\
\citet{ms-empirical-human} & - & - & - & 74.1 & 36.3 & 61.3 & 61.4 \\
\citet{choe-etal-2019-neural} & \textbf{76.2} & 50.3 & 69.1 & \textbf{74.8} & 34.1 & 60.3 & - \\
\citet{grundkiewicz-etal-2019-neural} & 72.3 & 60.1 & 69.5 & - & - & 64.2 & 61.2 \\
\citet{gec-pseudo-data} & 74.7 & 56.7 & 70.2 & 72.4 & \textbf{46.1} & \textbf{65.0} & 61.4 \\
\hline
\textbf{This work (ERRANT tags)} &  &  &  &  &  &  &  \\
5-Ensemble & 68.8 & \textbf{63.4} & 67.7 & 72.0 & 39.4 & 61.7 & 64.2 \\
+ Full sequence rescoring & 72.7 & 62.9 & \textbf{70.5} & 69.9 & 44.4 & 62.7 & \textbf{64.3} \\
\hline
\end{tabular}
\caption{\label{tab:gec-compare-ensemble} Ensemble results for grammatical error correction. Our full sequence baseline achieves 68.2 F$_{0.5}$ on BEA-test, 63.8 F$_{0.5}$  on CoNLL-14, and 62.4 GLEU on JFLEG-test.}
\end{table*}

\subsection{Grammatical Error Correction}

For grammatical error correction we follow a multi-stage fine-tuning setup.\footnote{Multi-stage fine-tuning has been proven effective for other sequence tasks such as machine translation \citep{khan-etal-2018-hunter,saunders-etal-2019-ucam}.} After training on the common pre-training data described above, we fine-tune for 30K steps on the public Lang-8~\citep{mizumoto-etal-2012-effect} corpus, followed by 500 steps on the FCE~\citep{yannakoudakis-etal-2011-new} and W\&I~\citep{bryant-etal-2019-bea} corpora. To improve comparability with related work across the different corpora we use ERRANT~\citep{bryant-etal-2017-automatic,felice-etal-2016-automatic} to compute span-based P(recision)$\uparrow$, R(ecall)$\uparrow$, and F$_{0.5}$-scores on the BEA development and test sets~\citep{bryant-etal-2019-bea}, the M2 scorer~\citep{dahlmeier-ng-2012-better} on the CoNLL-2014~\citep{ng-etal-2014-conll} test set, and GLEU$\uparrow$ on the JFLEG test set~\citep{napoles-etal-2017-jfleg}. However, the training data used in the literature varies vastly from system to system which limits comparability as (synthetic) data significantly impacts the system performance~\citep{grundkiewicz-etal-2019-neural}.

Table~\ref{tab:gec-compare-single-model} compares our approach with the best single model results reported in the literature. Our model tends to have a lower precision but higher recall than other systems. We are able to achieve the highest GLEU score on the JFLEG test set.

To further improve performance, we applied two techniques commonly used for grammatical error correction~\citep[inter alia]{grundkiewicz-etal-2019-neural}: ensembling and rescoring. Table~\ref{tab:gec-compare-ensemble} compares our 5-ensemble with other ensembles in the literature. Rescoring the $n$-best list from the edit model with a big full sequence Transformer model yields significant gains, outperforming all other systems in Table~\ref{tab:gec-compare-ensemble} on BEA-test and JFLEG.\footnote{This 
resembles the inverse setup of \citet{seq2seqforgec} who used edit features to rescore a full sequence model.}

\begin{table*}
\centering
\small
\begin{tabular}{lcccccc}
\hline \textbf{English} & \textbf{Avg.\ source} & \textbf{Avg.\ number} & \multicolumn{4}{c}{\textbf{Sentences per second$\uparrow$}} \\
\textbf{proficiency} & \textbf{length ($I$)} & \textbf{of edits ($N$)} & \textbf{Full} & \multicolumn{3}{c}{\textbf{Edit operation based}} \\
&  &  & \textbf{sequence} & \textbf{ERRANT tags}  & \textbf{No tags} & \textbf{ERRANT tags} \\
&  &  &  &   &  & \textbf{(with shortcuts)} \\ \hline
Beginner (CEFR-A) & 20.4 & 7.8 & 0.34 & 0.55 (1.6x) & 0.69 (2.0x)  & 0.69 (2.0x) \\
Intermediate (CEFR-B) & 25.9 & 6.0 & 0.34 & 0.83 (2.4x)  & 1.04 (3.0x) & 1.07 (3.1x) \\
Advanced (CEFR-C) & 23.0 & 4.2 & 0.31 & 1.21 (3.9x)  & 1.46 (4.7x) & 1.59 (\textbf{5.2x}) \\
Native & 26.6 & 4.0 & 0.26 & 1.03 (4.0x)  & 1.24 (4.8x) & 1.34 (\textbf{5.2x}) \\ \hline
\end{tabular}
\caption{\label{tab:speed-up} CPU decoding speeds without iterative refinement on BEA-dev averaged over three runs. Speed-ups compared to the full sequence baseline are in parentheses.}
\end{table*}

\begin{table*}
\centering
\small
\begin{tabular}{ccccccc}
\hline \multicolumn{2}{c}{\textbf{Oracle constraints}} & \multicolumn{2}{c}{\textbf{Text normalization (SER$\downarrow$)}} & \multicolumn{3}{c}{\textbf{Grammar correction (BEA-dev)}}  \\
\textbf{Tag ($t_n$)} & \textbf{Span end position ($p_n$)} & \textbf{English} & \textbf{Russian} & \textbf{P$\uparrow$} & \textbf{R$\uparrow$} & \textbf{F$_{0.5}\uparrow$} \\ \hline
& & 1.36 & 3.95 & 55.2 & 36.2 & 50.0 \\
$\checkmark$ &  & 0.36 & 3.63 & 58.6 & 53.5 & 57.5 \\
& $\checkmark$ & 0.48 & 3.58 & 64.9 & 65.5 & 65.0 \\
$\checkmark$ & $\checkmark$ & 0.24 & 3.58 & 71.9 & 71.9 & 71.9 \\
\hline
\end{tabular}
\caption{\label{tab:oracle-constraints} Partially constraining the decoder with oracle tags and/or span positions (no iterative refinement).}
\end{table*}

One of our initial goals was to avoid the wasteful computation of full sequence models when applied to tasks like grammatical error correction with a high degree of copying. Table~\ref{tab:speed-up} summarizes CPU decoding times on an Intel\textsuperscript{\tiny\textregistered} Xeon\textsuperscript{\tiny\textregistered} W-2135 Processor (12-core, 3.7 GHz).\footnote{We note that our experimental decoder implementation is not optimized for speed, and that absolute runtimes may differ with more efficient implementations.} We break down the measurements by English proficiency level. The full sequence baseline slows down for higher proficiency levels as sentences tend to be longer (second column of Table~\ref{tab:speed-up}). In contrast, our edit operation based approach is faster because it
does not depend on the target sequence length but instead on the number of edits which is usually small for advanced and native English speakers. We report speed-ups of 4.7-4.8x in these cases without using tags. When using tags, we implemented the following simple heuristics to improve the runtime (``shortcuts''): 1) If the tag $t_n=\mathtt{SELF}$ is predicted, directly set $r_n=\mathtt{SELF}$ and skip the replacement token prediction. 2) If the tag $t_n=\mathtt{EOS}$ is predicted, set $p_n=I$ and $r_n=\mathtt{EOS}$ and skip the span end position and the replacement token predictions.
These shortcuts do not affect the results in practice but provide speed-ups of 5.2x for advanced and native English compared to a full sequence model.

The speed-ups of our approach are mainly due to the shorter target sequence length compared to a full sequence model. However, our inference scheme in Sec.~\ref{sec:inference} still needs three times more time steps than the number of edits, i.e.\ around $4.0\times 3=12$ for native English (last row in Table \ref{tab:speed-up}). The observed speed-ups of 4.0x-5.2x are even larger than we would expect based on an average source length of $I=26.6$. One reason for the larger speed-ups is that the decoding runtime complexity under the Transformer is quadratic in length, not linear. Another reason is that although the three elements are predicted sequentially, not each prediction step is equally expensive: the softmax for the tag and span predictions is computed over only a couple of elements, not over the full 32K subword vocabulary. Furthermore, efficient decoder implementations could reuse most of the computation done for the tag prediction for the span position.

\subsection{Oracle Experiments}

Our model can be viewed from a multi-task perspective since it tries to predict three different features (tag, span position, and replacement token). To better understand the contributions of these different components we partially constrained them using the gold references for both text normalization and grammatical error correction tasks. We avoid constraining the {\em number} of edits ($N$) in these oracle experiments by giving the constrained decoder the option to repeat labels in the reference. Table~\ref{tab:oracle-constraints} shows that having access to the gold tags and/or span positions greatly improves performance. We hypothesize that these gains can be largely attributed to the resolution of confusion between self and non-self. An interesting outlier is text normalization on Russian which benefits less from oracle constraints. This suggests that the challenges for Russian text normalization are largely in predicting the replacement tokens, possibly due to the morphological complexity of Russian.

\begin{table}
\centering
\small
\begin{tabular}{lccc}
\hline \textbf{System} & \multicolumn{3}{c}{\textbf{Tagging accuracy}} \\
& \textbf{P$\uparrow$} & \textbf{R$\uparrow$}  & \textbf{F$_{0.5}\uparrow$} \\ \hline
Lasertagger & 54.9 & 33.7 & 48.8 \\
This work (unconstrained) & 67.9 & 25.8 & 51.2 \\
This work (span-constrained) & 94.7 & 52.4 & 81.5 \\ \hline
\end{tabular}
\caption{\label{tab:tag-accuracy} Tagging accuracy on BEA-dev (no iterative refinement).}
\end{table}

Since the motivation for using tags was to improve explainability we also evaluated the accuracy of the tag prediction on grammatical error correction. For comparison, we trained a baseline Lasertagger~\citep{lasertagger} on a subset of the BEA training set (30.4K examples) to predict the ERRANT tags. Insertions are represented as composite tags together with the subsequent tag such that the total Lasertagger vocabulary size is 213.
The model was initialized from a pre-trained BERT~\citep{devlin-etal-2019-bert} checkpoint. Decoding was performed using an autoregressive strategy with a Transformer decoder. We used the default hyper-parameters without any task-specific optimization.
Table~\ref{tab:tag-accuracy} shows that the tag prediction of our unconstrained model is more accurate than the Lasertagger baseline. Errors in this unconstrained setup are either due to predicting the wrong tag or predicting the wrong span. To tease apart these error sources we also report the accuracy under oracle span constraints. Our span constrained model achieves a recall of 52.4\%, i.e.\ more than half of the non-self tags are classified correctly (28 tags).

\section{Related Work}
A popular way to tackle NLP problems with overlap between input and output is to equip seq2seq models with a copying mechanism~\citep{jia-liang-2016-data,zhao-etal-2019-improving,chen-bansal-2018-fast,nallapati-etal-2016-abstractive,gulcehre-etal-2016-pointing,see-etal-2017-get,gu-etal-2016-incorporating}, usually borrowing ideas from pointer networks~\citep{pointer-networks} to point to single tokens in the source sequence. In contrast, we use pointer networks to identify entire spans that are to be copied which results in a much more compact representation and faster decoding. The idea of using span-level edits has also been explored for morphological learning in the form of symbolic span-level rules, but not in combination with neural models \citep{elsner2019modeling}.

Our work is related to neural multi-task learning for NLP~\citep{uni-arch-multitask,dong-etal-2015-multi,multitask-seq2seq,sogaard-goldberg-2016-deep}. Unlike multi-task learning which typically solves separate problems (e.g.\ POS tagging and named entity recognition~\citep{uni-arch-multitask} or translation into different languages~\citep{multitask-seq2seq,dong-etal-2015-multi}) with the same model, our three output features (tag, source span, and replacement) represent the same output sequence (Algorithm ~\ref{alg:edit2seq}). Thus, it resembles the stack-propagation approach of \citet{zhang-weiss-2016-stack} who use POS tags to improve parsing performance.  

A more recent line of research frames sequence editing as a labelling problem using labels such as \texttt{ADD}, \texttt{KEEP}, and \texttt{DELETE}~\citep{ribeiro-etal-2018-local,dong-etal-2019-editnts,felix,lasertagger,edit-tag}, often relying heavily on BERT~\citep{devlin-etal-2019-bert} pre-training. Similar operations such as insertions and deletions have also been used for machine translation~\citep{lev-transformer,insertion-transformer,insertion-decoding,insertion-low-resource,stahlberg-etal-2018-operation}. We showed in Sec.~\ref{sec:experiments} that our model often performs similarly or better than those approaches, with the added advantage of providing explanations for its predictions. 

\section{Discussion}
We have presented a neural model that represents sequence transduction using span-based edit operations. We reported competitive results on five different NLP problems, improving the state of the art on text normalization, sentence splitting, and the JFLEG test set for grammatical error correction. We showed that our approach is 2.0-5.2 times faster than a full sequence model for grammatical error correction. Our model can predict labels that explain each edit to improve the interpretability for the end-user. However, we do not make any claim that \emph{Seq2Edits} can provide insights into the internal mechanics of the neural model. The underlying neural model in \emph{Seq2Edits} is as much of a black-box as a regular full sequence model. 

While our model is advantageous in terms of speed and explainability, it does have some weaknesses. Notably, the model uses a tailored architecture (Figure~\ref{fig:architecture}) that would require some engineering effort to implement efficiently. Second, the output of the model tends to be less fluent than a regular full sequence model, as can be seen from the examples in Table~\ref{tab:examples-gec}. This is not an issue for localized edit tasks such as text normalization but may be a drawback for tasks involving substantial rewrites (e.g. GEC for non-native speakers).

Even though our approach is open-vocabulary, future work will explore task specific restrictions. For example, in a model for dialog applications, we may want to restrict the set of response tokens to a predefined list. Alternatively, it may be useful to explore generation in a non left-to-right order to improve the efficiency of inference.

Another line of future work is to extend our model to sequence rewriting tasks, such as Machine Translation post-editing, that do not have existing error-tag dictionaries. This research would require induction of error tag inventories using either linguistic insights or unsupervised methods.

\section*{Acknowledgments}

We thank Chris Alberti and Jared Lichtarge as well as the anonymous reviewers for their helpful comments.

\bibliographystyle{acl_natbib}
\bibliography{anthology,emnlp2020}

\begin{thebibliography}{60}
\expandafter\ifx\csname natexlab\endcsname\relax\def\natexlab#1{#1}\fi

\bibitem[{Awasthi et~al.(2019)Awasthi, Sarawagi, Goyal, Ghosh, and
  Piratla}]{edit-tag}
Abhijeet Awasthi, Sunita Sarawagi, Rasna Goyal, Sabyasachi Ghosh, and Vihari
  Piratla. 2019.
\newblock \href {https://doi.org/10.18653/v1/D19-1435} {Parallel iterative edit
  models for local sequence transduction}.
\newblock In \emph{Proceedings of the 2019 Conference on Empirical Methods in
  Natural Language Processing and the 9th International Joint Conference on
  Natural Language Processing (EMNLP-IJCNLP)}, pages 4260--4270, Hong Kong,
  China. Association for Computational Linguistics.

\bibitem[{Bahdanau et~al.(2015)Bahdanau, Cho, and Bengio}]{bahdanau-nmt}
Dzmitry Bahdanau, {Kyung Hyun} Cho, and Yoshua Bengio. 2015.
\newblock Neural machine translation by jointly learning to align and
  translate.
\newblock In \emph{3rd International Conference on Learning Representations,
  ICLR 2015}.

\bibitem[{Botha et~al.(2018)Botha, Faruqui, Alex, Baldridge, and
  Das}]{botha-etal-2018-learning}
Jan~A. Botha, Manaal Faruqui, John Alex, Jason Baldridge, and Dipanjan Das.
  2018.
\newblock \href {https://doi.org/10.18653/v1/D18-1080} {Learning to split and
  rephrase from {W}ikipedia edit history}.
\newblock In \emph{Proceedings of the 2018 Conference on Empirical Methods in
  Natural Language Processing}, pages 732--737, Brussels, Belgium. Association
  for Computational Linguistics.

\bibitem[{Bryant et~al.(2019)Bryant, Felice, Andersen, and
  Briscoe}]{bryant-etal-2019-bea}
Christopher Bryant, Mariano Felice, {\O}istein~E. Andersen, and Ted Briscoe.
  2019.
\newblock \href {https://doi.org/10.18653/v1/W19-4406} {The {BEA}-2019 shared
  task on grammatical error correction}.
\newblock In \emph{Proceedings of the Fourteenth Workshop on Innovative Use of
  NLP for Building Educational Applications}, pages 52--75, Florence, Italy.
  Association for Computational Linguistics.

\bibitem[{Bryant et~al.(2017)Bryant, Felice, and
  Briscoe}]{bryant-etal-2017-automatic}
Christopher Bryant, Mariano Felice, and Ted Briscoe. 2017.
\newblock \href {https://doi.org/10.18653/v1/P17-1074} {Automatic annotation
  and evaluation of error types for grammatical error correction}.
\newblock In \emph{Proceedings of the 55th Annual Meeting of the Association
  for Computational Linguistics (Volume 1: Long Papers)}, pages 793--805,
  Vancouver, Canada. Association for Computational Linguistics.

\bibitem[{Chen and Bansal(2018)}]{chen-bansal-2018-fast}
Yen-Chun Chen and Mohit Bansal. 2018.
\newblock \href {https://doi.org/10.18653/v1/P18-1063} {Fast abstractive
  summarization with reinforce-selected sentence rewriting}.
\newblock In \emph{Proceedings of the 56th Annual Meeting of the Association
  for Computational Linguistics (Volume 1: Long Papers)}, pages 675--686,
  Melbourne, Australia. Association for Computational Linguistics.

\bibitem[{Choe et~al.(2019)Choe, Ham, Park, and Yoon}]{choe-etal-2019-neural}
Yo~Joong Choe, Jiyeon Ham, Kyubyong Park, and Yeoil Yoon. 2019.
\newblock \href {https://doi.org/10.18653/v1/W19-4423} {A neural grammatical
  error correction system built on better pre-training and sequential transfer
  learning}.
\newblock In \emph{Proceedings of the Fourteenth Workshop on Innovative Use of
  NLP for Building Educational Applications}, pages 213--227, Florence, Italy.
  Association for Computational Linguistics.

\bibitem[{Chollampatt and Ng(2018)}]{seq2seqforgec}
Shamil Chollampatt and Hwee~Tou Ng. 2018.
\newblock A multilayer convolutional encoder-decoder neural network for
  grammatical error correction.
\newblock In \emph{Thirty-Second AAAI Conference on Artificial Intelligence}.

\bibitem[{Collobert and Weston(2008)}]{uni-arch-multitask}
Ronan Collobert and Jason Weston. 2008.
\newblock \href {https://doi.org/10.1145/1390156.1390177} {A unified
  architecture for natural language processing: Deep neural networks with
  multitask learning}.
\newblock In \emph{Proceedings of the 25th International Conference on Machine
  Learning}, ICML ’08, page 160–167, New York, NY, USA. Association for
  Computing Machinery.

\bibitem[{Dahlmeier and Ng(2012)}]{dahlmeier-ng-2012-better}
Daniel Dahlmeier and Hwee~Tou Ng. 2012.
\newblock \href {https://www.aclweb.org/anthology/N12-1067} {Better evaluation
  for grammatical error correction}.
\newblock In \emph{Proceedings of the 2012 Conference of the North {A}merican
  Chapter of the Association for Computational Linguistics: Human Language
  Technologies}, pages 568--572, Montr{\'e}al, Canada. Association for
  Computational Linguistics.

\bibitem[{Devlin et~al.(2019)Devlin, Chang, Lee, and
  Toutanova}]{devlin-etal-2019-bert}
Jacob Devlin, Ming-Wei Chang, Kenton Lee, and Kristina Toutanova. 2019.
\newblock \href {https://doi.org/10.18653/v1/N19-1423} {{BERT}: Pre-training of
  deep bidirectional transformers for language understanding}.
\newblock In \emph{Proceedings of the 2019 Conference of the North {A}merican
  Chapter of the Association for Computational Linguistics: Human Language
  Technologies, Volume 1 (Long and Short Papers)}, pages 4171--4186,
  Minneapolis, Minnesota. Association for Computational Linguistics.

\bibitem[{Dong et~al.(2015)Dong, Wu, He, Yu, and Wang}]{dong-etal-2015-multi}
Daxiang Dong, Hua Wu, Wei He, Dianhai Yu, and Haifeng Wang. 2015.
\newblock \href {https://doi.org/10.3115/v1/P15-1166} {Multi-task learning for
  multiple language translation}.
\newblock In \emph{Proceedings of the 53rd Annual Meeting of the Association
  for Computational Linguistics and the 7th International Joint Conference on
  Natural Language Processing (Volume 1: Long Papers)}, pages 1723--1732,
  Beijing, China. Association for Computational Linguistics.

\bibitem[{Dong et~al.(2019)Dong, Li, Rezagholizadeh, and
  Cheung}]{dong-etal-2019-editnts}
Yue Dong, Zichao Li, Mehdi Rezagholizadeh, and Jackie Chi~Kit Cheung. 2019.
\newblock \href {https://doi.org/10.18653/v1/P19-1331} {{E}dit{NTS}: An neural
  programmer-interpreter model for sentence simplification through explicit
  editing}.
\newblock In \emph{Proceedings of the 57th Annual Meeting of the Association
  for Computational Linguistics}, pages 3393--3402, Florence, Italy.
  Association for Computational Linguistics.

\bibitem[{Elsner et~al.(2019)Elsner, Sims, Erdmann, Hernandez, Jaffe, Jin,
  Johnson, Karim, King, Nunes et~al.}]{elsner2019modeling}
Micha Elsner, Andrea~D Sims, Alexander Erdmann, Antonio Hernandez, Evan Jaffe,
  Lifeng Jin, Martha~Booker Johnson, Shuan Karim, David~L King, Luana~Lamberti
  Nunes, et~al. 2019.
\newblock \href
  {http://jlm.ipipan.waw.pl/ojs/index.php/JLM/article/viewFile/244/238}
  {Modeling morphological learning, typology, and change: What can the neural
  sequence-to-sequence framework contribute?}
\newblock \emph{Journal of Language Modelling}, 7(1):53--98.

\bibitem[{Felice et~al.(2016)Felice, Bryant, and
  Briscoe}]{felice-etal-2016-automatic}
Mariano Felice, Christopher Bryant, and Ted Briscoe. 2016.
\newblock \href {https://www.aclweb.org/anthology/C16-1079} {Automatic
  extraction of learner errors in {ESL} sentences using linguistically enhanced
  alignments}.
\newblock In \emph{Proceedings of {COLING} 2016, the 26th International
  Conference on Computational Linguistics: Technical Papers}, pages 825--835,
  Osaka, Japan. The COLING 2016 Organizing Committee.

\bibitem[{Ge et~al.(2018)Ge, Wei, and Zhou}]{ms-empirical-human}
Tao Ge, Furu Wei, and Ming Zhou. 2018.
\newblock Reaching human-level performance in automatic grammatical error
  correction: An empirical study.
\newblock \emph{arXiv preprint arXiv:1807.01270}.

\bibitem[{Geva et~al.(2019)Geva, Malmi, Szpektor, and
  Berant}]{geva-etal-2019-discofuse}
Mor Geva, Eric Malmi, Idan Szpektor, and Jonathan Berant. 2019.
\newblock \href {https://doi.org/10.18653/v1/N19-1348} {{D}isco{F}use: A
  large-scale dataset for discourse-based sentence fusion}.
\newblock In \emph{Proceedings of the 2019 Conference of the North {A}merican
  Chapter of the Association for Computational Linguistics: Human Language
  Technologies, Volume 1 (Long and Short Papers)}, pages 3443--3455,
  Minneapolis, Minnesota. Association for Computational Linguistics.

\bibitem[{Grundkiewicz et~al.(2019)Grundkiewicz, Junczys-Dowmunt, and
  Heafield}]{grundkiewicz-etal-2019-neural}
Roman Grundkiewicz, Marcin Junczys-Dowmunt, and Kenneth Heafield. 2019.
\newblock \href {https://doi.org/10.18653/v1/W19-4427} {Neural grammatical
  error correction systems with unsupervised pre-training on synthetic data}.
\newblock In \emph{Proceedings of the Fourteenth Workshop on Innovative Use of
  NLP for Building Educational Applications}, pages 252--263, Florence, Italy.
  Association for Computational Linguistics.

\bibitem[{Gu et~al.(2019{\natexlab{a}})Gu, Liu, and Cho}]{insertion-decoding}
Jiatao Gu, Qi~Liu, and Kyunghyun Cho. 2019{\natexlab{a}}.
\newblock \href {https://doi.org/10.1162/tacl\_a\_00292} {Insertion-based
  decoding with automatically inferred generation order}.
\newblock \emph{Transactions of the Association for Computational Linguistics},
  7:661--676.

\bibitem[{Gu et~al.(2016)Gu, Lu, Li, and Li}]{gu-etal-2016-incorporating}
Jiatao Gu, Zhengdong Lu, Hang Li, and Victor~O.K. Li. 2016.
\newblock \href {https://doi.org/10.18653/v1/P16-1154} {Incorporating copying
  mechanism in sequence-to-sequence learning}.
\newblock In \emph{Proceedings of the 54th Annual Meeting of the Association
  for Computational Linguistics (Volume 1: Long Papers)}, pages 1631--1640,
  Berlin, Germany. Association for Computational Linguistics.

\bibitem[{Gu et~al.(2019{\natexlab{b}})Gu, Wang, and Zhao}]{lev-transformer}
Jiatao Gu, Changhan Wang, and Junbo Zhao. 2019{\natexlab{b}}.
\newblock \href {http://papers.nips.cc/paper/9297-levenshtein-transformer.pdf}
  {Levenshtein transformer}.
\newblock In H.~Wallach, H.~Larochelle, A.~Beygelzimer, F.~d'Alch\'{e} Buc,
  E.~Fox, and R.~Garnett, editors, \emph{Advances in Neural Information
  Processing Systems 32}, pages 11181--11191. Curran Associates, Inc.

\bibitem[{Gulcehre et~al.(2016)Gulcehre, Ahn, Nallapati, Zhou, and
  Bengio}]{gulcehre-etal-2016-pointing}
Caglar Gulcehre, Sungjin Ahn, Ramesh Nallapati, Bowen Zhou, and Yoshua Bengio.
  2016.
\newblock \href {https://doi.org/10.18653/v1/P16-1014} {Pointing the unknown
  words}.
\newblock In \emph{Proceedings of the 54th Annual Meeting of the Association
  for Computational Linguistics (Volume 1: Long Papers)}, pages 140--149,
  Berlin, Germany. Association for Computational Linguistics.

\bibitem[{He et~al.(2016)He, Zhang, Ren, and Sun}]{residual-connections}
Kaiming He, Xiangyu Zhang, Shaoqing Ren, and Jian Sun. 2016.
\newblock Deep residual learning for image recognition.
\newblock In \emph{Proceedings of the IEEE conference on computer vision and
  pattern recognition}, pages 770--778.

\bibitem[{Jia and Liang(2016)}]{jia-liang-2016-data}
Robin Jia and Percy Liang. 2016.
\newblock \href {https://doi.org/10.18653/v1/P16-1002} {Data recombination for
  neural semantic parsing}.
\newblock In \emph{Proceedings of the 54th Annual Meeting of the Association
  for Computational Linguistics (Volume 1: Long Papers)}, pages 12--22, Berlin,
  Germany. Association for Computational Linguistics.

\bibitem[{Kalchbrenner and
  Blunsom(2013)}]{kalchbrenner-blunsom-2013-recurrent-continuous}
Nal Kalchbrenner and Phil Blunsom. 2013.
\newblock \href {https://www.aclweb.org/anthology/D13-1176} {Recurrent
  continuous translation models}.
\newblock In \emph{Proceedings of the 2013 Conference on Empirical Methods in
  Natural Language Processing}, pages 1700--1709, Seattle, Washington, USA.
  Association for Computational Linguistics.

\bibitem[{Khan et~al.(2018)Khan, Panda, Xu, and Flokas}]{khan-etal-2018-hunter}
Abdul Khan, Subhadarshi Panda, Jia Xu, and Lampros Flokas. 2018.
\newblock \href {https://doi.org/10.18653/v1/W18-6447} {Hunter {NMT} system for
  {WMT}18 biomedical translation task: Transfer learning in neural machine
  translation}.
\newblock In \emph{Proceedings of the Third Conference on Machine Translation:
  Shared Task Papers}, pages 655--661, Belgium, Brussels. Association for
  Computational Linguistics.

\bibitem[{Kiyono et~al.(2019)Kiyono, Suzuki, Mita, Mizumoto, and
  Inui}]{gec-pseudo-data}
Shun Kiyono, Jun Suzuki, Masato Mita, Tomoya Mizumoto, and Kentaro Inui. 2019.
\newblock \href {https://doi.org/10.18653/v1/D19-1119} {An empirical study of
  incorporating pseudo data into grammatical error correction}.
\newblock In \emph{Proceedings of the 2019 Conference on Empirical Methods in
  Natural Language Processing and the 9th International Joint Conference on
  Natural Language Processing (EMNLP-IJCNLP)}, pages 1236--1242, Hong Kong,
  China. Association for Computational Linguistics.

\bibitem[{Lichtarge et~al.(2019)Lichtarge, Alberti, Kumar, Shazeer, Parmar, and
  Tong}]{lichtarge-etal-2019-corpora}
Jared Lichtarge, Chris Alberti, Shankar Kumar, Noam Shazeer, Niki Parmar, and
  Simon Tong. 2019.
\newblock \href {https://doi.org/10.18653/v1/N19-1333} {Corpora generation for
  grammatical error correction}.
\newblock In \emph{Proceedings of the 2019 Conference of the North {A}merican
  Chapter of the Association for Computational Linguistics: Human Language
  Technologies, Volume 1 (Long and Short Papers)}, pages 3291--3301,
  Minneapolis, Minnesota. Association for Computational Linguistics.

\bibitem[{Luong et~al.(2015)Luong, Le, Sutskever, Vinyals, and
  Kaiser}]{multitask-seq2seq}
Minh-Thang Luong, Quoc~V Le, Ilya Sutskever, Oriol Vinyals, and Lukasz Kaiser.
  2015.
\newblock Multi-task sequence to sequence learning.
\newblock \emph{arXiv preprint arXiv:1511.06114}.

\bibitem[{Mallinson et~al.(2020)Mallinson, Severyn, Malmi, and Garrido}]{felix}
Jonathan Mallinson, Aliaksei Severyn, Eric Malmi, and Guillermo Garrido. 2020.
\newblock Felix: Flexible text editing through tagging and insertion.
\newblock \emph{arXiv preprint arXiv:2003.10687}.

\bibitem[{Malmi et~al.(2019)Malmi, Krause, Rothe, Mirylenka, and
  Severyn}]{lasertagger}
Eric Malmi, Sebastian Krause, Sascha Rothe, Daniil Mirylenka, and Aliaksei
  Severyn. 2019.
\newblock \href {https://doi.org/10.18653/v1/D19-1510} {Encode, tag, realize:
  High-precision text editing}.
\newblock In \emph{Proceedings of the 2019 Conference on Empirical Methods in
  Natural Language Processing and the 9th International Joint Conference on
  Natural Language Processing (EMNLP-IJCNLP)}, pages 5054--5065, Hong Kong,
  China. Association for Computational Linguistics.

\bibitem[{Mansfield et~al.(2019)Mansfield, Sun, Liu, Gandhe, and
  Hoffmeister}]{mansfield-etal-2019-neural}
Courtney Mansfield, Ming Sun, Yuzong Liu, Ankur Gandhe, and Bj{\"o}rn
  Hoffmeister. 2019.
\newblock \href {https://doi.org/10.18653/v1/N19-2024} {Neural text
  normalization with subword units}.
\newblock In \emph{Proceedings of the 2019 Conference of the North {A}merican
  Chapter of the Association for Computational Linguistics: Human Language
  Technologies, Volume 2 (Industry Papers)}, pages 190--196, Minneapolis -
  Minnesota. Association for Computational Linguistics.

\bibitem[{Mizumoto et~al.(2012)Mizumoto, Hayashibe, Komachi, Nagata, and
  Matsumoto}]{mizumoto-etal-2012-effect}
Tomoya Mizumoto, Yuta Hayashibe, Mamoru Komachi, Masaaki Nagata, and Yuji
  Matsumoto. 2012.
\newblock \href {https://www.aclweb.org/anthology/C12-2084} {The effect of
  learner corpus size in grammatical error correction of {ESL} writings}.
\newblock In \emph{Proceedings of {COLING} 2012: Posters}, pages 863--872,
  Mumbai, India. The COLING 2012 Organizing Committee.

\bibitem[{Nallapati et~al.(2016)Nallapati, Zhou, dos Santos, Gu̇l{\c{c}}ehre,
  and Xiang}]{nallapati-etal-2016-abstractive}
Ramesh Nallapati, Bowen Zhou, Cicero dos Santos, {\c{C}}a{\u{g}}lar
  Gu̇l{\c{c}}ehre, and Bing Xiang. 2016.
\newblock \href {https://doi.org/10.18653/v1/K16-1028} {Abstractive text
  summarization using sequence-to-sequence {RNN}s and beyond}.
\newblock In \emph{Proceedings of The 20th {SIGNLL} Conference on Computational
  Natural Language Learning}, pages 280--290, Berlin, Germany. Association for
  Computational Linguistics.

\bibitem[{Napoles et~al.(2017)Napoles, Sakaguchi, and
  Tetreault}]{napoles-etal-2017-jfleg}
Courtney Napoles, Keisuke Sakaguchi, and Joel Tetreault. 2017.
\newblock \href {https://www.aclweb.org/anthology/E17-2037} {{JFLEG}: A fluency
  corpus and benchmark for grammatical error correction}.
\newblock In \emph{Proceedings of the 15th Conference of the {E}uropean Chapter
  of the Association for Computational Linguistics: Volume 2, Short Papers},
  pages 229--234, Valencia, Spain. Association for Computational Linguistics.

\bibitem[{Ng et~al.(2014)Ng, Wu, Briscoe, Hadiwinoto, Susanto, and
  Bryant}]{ng-etal-2014-conll}
Hwee~Tou Ng, Siew~Mei Wu, Ted Briscoe, Christian Hadiwinoto, Raymond~Hendy
  Susanto, and Christopher Bryant. 2014.
\newblock \href {https://doi.org/10.3115/v1/W14-1701} {The {C}o{NLL}-2014
  shared task on grammatical error correction}.
\newblock In \emph{Proceedings of the Eighteenth Conference on Computational
  Natural Language Learning: Shared Task}, pages 1--14, Baltimore, Maryland.
  Association for Computational Linguistics.

\bibitem[{{\"O}stling and Tiedemann(2017)}]{insertion-low-resource}
Robert {\"O}stling and J{\"o}rg Tiedemann. 2017.
\newblock Neural machine translation for low-resource languages.
\newblock \emph{arXiv preprint arXiv:1708.05729}.

\bibitem[{Raffel et~al.(2019)Raffel, Shazeer, Roberts, Lee, Narang, Matena,
  Zhou, Li, and Liu}]{t5}
Colin Raffel, Noam Shazeer, Adam Roberts, Katherine Lee, Sharan Narang, Michael
  Matena, Yanqi Zhou, Wei Li, and Peter~J Liu. 2019.
\newblock Exploring the limits of transfer learning with a unified text-to-text
  transformer.
\newblock \emph{arXiv preprint arXiv:1910.10683}.

\bibitem[{Ribeiro et~al.(2018)Ribeiro, Narayan, Cohen, and
  Carreras}]{ribeiro-etal-2018-local}
Joana Ribeiro, Shashi Narayan, Shay~B. Cohen, and Xavier Carreras. 2018.
\newblock \href {https://www.aclweb.org/anthology/C18-1115} {Local string
  transduction as sequence labeling}.
\newblock In \emph{Proceedings of the 27th International Conference on
  Computational Linguistics}, pages 1360--1371, Santa Fe, New Mexico, USA.
  Association for Computational Linguistics.

\bibitem[{Rothe et~al.(2019)Rothe, Narayan, and Severyn}]{bert2bert}
Sascha Rothe, Shashi Narayan, and Aliaksei Severyn. 2019.
\newblock Leveraging pre-trained checkpoints for sequence generation tasks.
\newblock \emph{arXiv preprint arXiv:1907.12461}.

\bibitem[{Saunders et~al.(2019)Saunders, Stahlberg, and
  Byrne}]{saunders-etal-2019-ucam}
Danielle Saunders, Felix Stahlberg, and Bill Byrne. 2019.
\newblock \href {https://doi.org/10.18653/v1/W19-5421} {{UCAM} biomedical
  translation at {WMT}19: Transfer learning multi-domain ensembles}.
\newblock In \emph{Proceedings of the Fourth Conference on Machine Translation
  (Volume 3: Shared Task Papers, Day 2)}, pages 169--174, Florence, Italy.
  Association for Computational Linguistics.

\bibitem[{See et~al.(2017)See, Liu, and Manning}]{see-etal-2017-get}
Abigail See, Peter~J. Liu, and Christopher~D. Manning. 2017.
\newblock \href {https://doi.org/10.18653/v1/P17-1099} {Get to the point:
  Summarization with pointer-generator networks}.
\newblock In \emph{Proceedings of the 55th Annual Meeting of the Association
  for Computational Linguistics (Volume 1: Long Papers)}, pages 1073--1083,
  Vancouver, Canada. Association for Computational Linguistics.

\bibitem[{Sennrich et~al.(2016)Sennrich, Haddow, and
  Birch}]{sennrich-etal-2016-neural}
Rico Sennrich, Barry Haddow, and Alexandra Birch. 2016.
\newblock \href {https://doi.org/10.18653/v1/P16-1162} {Neural machine
  translation of rare words with subword units}.
\newblock In \emph{Proceedings of the 54th Annual Meeting of the Association
  for Computational Linguistics (Volume 1: Long Papers)}, pages 1715--1725,
  Berlin, Germany. Association for Computational Linguistics.

\bibitem[{Shazeer and Stern(2018)}]{adafactor}
Noam Shazeer and Mitchell Stern. 2018.
\newblock \href {http://proceedings.mlr.press/v80/shazeer18a.html} {Adafactor:
  Adaptive learning rates with sublinear memory cost}.
\newblock In \emph{Proceedings of the 35th International Conference on Machine
  Learning}, volume~80 of \emph{Proceedings of Machine Learning Research},
  pages 4596--4604, Stockholmsmässan, Stockholm Sweden. PMLR.

\bibitem[{S{\o}gaard and Goldberg(2016)}]{sogaard-goldberg-2016-deep}
Anders S{\o}gaard and Yoav Goldberg. 2016.
\newblock \href {https://doi.org/10.18653/v1/P16-2038} {Deep multi-task
  learning with low level tasks supervised at lower layers}.
\newblock In \emph{Proceedings of the 54th Annual Meeting of the Association
  for Computational Linguistics (Volume 2: Short Papers)}, pages 231--235,
  Berlin, Germany. Association for Computational Linguistics.

\bibitem[{Sproat and Jaitly(2016)}]{rws-rnn-text-norm}
Richard Sproat and Navdeep Jaitly. 2016.
\newblock {RNN} approaches to text normalization: {A} challenge.
\newblock \emph{arXiv preprint arXiv:1611.00068}.

\bibitem[{Stahlberg et~al.(2018)Stahlberg, Saunders, and
  Byrne}]{stahlberg-etal-2018-operation}
Felix Stahlberg, Danielle Saunders, and Bill Byrne. 2018.
\newblock \href {https://doi.org/10.18653/v1/W18-5420} {An operation sequence
  model for explainable neural machine translation}.
\newblock In \emph{Proceedings of the 2018 {EMNLP} Workshop {B}lackbox{NLP}:
  Analyzing and Interpreting Neural Networks for {NLP}}, pages 175--186,
  Brussels, Belgium. Association for Computational Linguistics.

\bibitem[{Stern et~al.(2019)Stern, Chan, Kiros, and
  Uszkoreit}]{insertion-transformer}
Mitchell Stern, William Chan, Jamie Kiros, and Jakob Uszkoreit. 2019.
\newblock \href {http://proceedings.mlr.press/v97/stern19a.html} {Insertion
  transformer: Flexible sequence generation via insertion operations}.
\newblock In \emph{Proceedings of the 36th International Conference on Machine
  Learning}, volume~97 of \emph{Proceedings of Machine Learning Research},
  pages 5976--5985, Long Beach, California, USA. PMLR.

\bibitem[{Sutskever et~al.(2014)Sutskever, Vinyals, and Le}]{sutskever-nmt}
Ilya Sutskever, Oriol Vinyals, and Quoc~V Le. 2014.
\newblock \href
  {http://papers.nips.cc/paper/5346-sequence-to-sequence-learning-with-neural-networks.pdf}
  {Sequence to sequence learning with neural networks}.
\newblock In Z.~Ghahramani, M.~Welling, C.~Cortes, N.~D. Lawrence, and K.~Q.
  Weinberger, editors, \emph{Advances in Neural Information Processing Systems
  27}, pages 3104--3112. Curran Associates, Inc.

\bibitem[{Tan et~al.(2017)Tan, Wan, and Xiao}]{seq2seqforsummarization}
Jiwei Tan, Xiaojun Wan, and Jianguo Xiao. 2017.
\newblock \href {https://doi.org/10.18653/v1/P17-1108} {Abstractive document
  summarization with a graph-based attentional neural model}.
\newblock In \emph{Proceedings of the 55th Annual Meeting of the Association
  for Computational Linguistics}, pages 1171--1181, Vancouver, Canada.
  Association for Computational Linguistics.

\bibitem[{Vaswani et~al.(2018)Vaswani, Bengio, Brevdo, Chollet, Gomez, Gouws,
  Jones, Kaiser, Kalchbrenner, Parmar, Sepassi, Shazeer, and
  Uszkoreit}]{vaswani-etal-2018-tensor2tensor}
Ashish Vaswani, Samy Bengio, Eugene Brevdo, Francois Chollet, Aidan Gomez,
  Stephan Gouws, Llion Jones, {\L}ukasz Kaiser, Nal Kalchbrenner, Niki Parmar,
  Ryan Sepassi, Noam Shazeer, and Jakob Uszkoreit. 2018.
\newblock \href {https://www.aclweb.org/anthology/W18-1819} {{T}ensor2{T}ensor
  for neural machine translation}.
\newblock In \emph{Proceedings of the 13th Conference of the Association for
  Machine Translation in the {A}mericas (Volume 1: Research Papers)}, pages
  193--199, Boston, MA. Association for Machine Translation in the Americas.

\bibitem[{Vaswani et~al.(2017)Vaswani, Shazeer, Parmar, Uszkoreit, Jones,
  Gomez, Kaiser, and Polosukhin}]{transformer}
Ashish Vaswani, Noam Shazeer, Niki Parmar, Jakob Uszkoreit, Llion Jones,
  Aidan~N Gomez, \L~ukasz Kaiser, and Illia Polosukhin. 2017.
\newblock \href
  {http://papers.nips.cc/paper/7181-attention-is-all-you-need.pdf} {Attention
  is all you need}.
\newblock In I.~Guyon, U.~V. Luxburg, S.~Bengio, H.~Wallach, R.~Fergus,
  S.~Vishwanathan, and R.~Garnett, editors, \emph{Advances in Neural
  Information Processing Systems 30}, pages 5998--6008. Curran Associates, Inc.

\bibitem[{Vinyals et~al.(2015)Vinyals, Fortunato, and
  Jaitly}]{pointer-networks}
Oriol Vinyals, Meire Fortunato, and Navdeep Jaitly. 2015.
\newblock \href {http://papers.nips.cc/paper/5866-pointer-networks.pdf}
  {Pointer networks}.
\newblock In C.~Cortes, N.~D. Lawrence, D.~D. Lee, M.~Sugiyama, and R.~Garnett,
  editors, \emph{Advances in Neural Information Processing Systems 28}, pages
  2692--2700. Curran Associates, Inc.

\bibitem[{Wu et~al.(2016)Wu, Schuster, Chen, Le, Norouzi, Macherey, Krikun,
  Cao, Gao, Macherey et~al.}]{gnmt}
Yonghui Wu, Mike Schuster, Zhifeng Chen, Quoc~V Le, Mohammad Norouzi, Wolfgang
  Macherey, Maxim Krikun, Yuan Cao, Qin Gao, Klaus Macherey, et~al. 2016.
\newblock Google's neural machine translation system: Bridging the gap between
  human and machine translation.
\newblock \emph{arXiv preprint arXiv:1609.08144}.

\bibitem[{Xu et~al.(2016)Xu, Napoles, Pavlick, Chen, and
  Callison-Burch}]{xu-etal-2016-optimizing}
Wei Xu, Courtney Napoles, Ellie Pavlick, Quanze Chen, and Chris Callison-Burch.
  2016.
\newblock \href {https://doi.org/10.1162/tacl_a_00107} {Optimizing statistical
  machine translation for text simplification}.
\newblock \emph{Transactions of the Association for Computational Linguistics},
  4:401--415.

\bibitem[{Yannakoudakis et~al.(2011)Yannakoudakis, Briscoe, and
  Medlock}]{yannakoudakis-etal-2011-new}
Helen Yannakoudakis, Ted Briscoe, and Ben Medlock. 2011.
\newblock \href {https://www.aclweb.org/anthology/P11-1019} {A new dataset and
  method for automatically grading {ESOL} texts}.
\newblock In \emph{Proceedings of the 49th Annual Meeting of the Association
  for Computational Linguistics: Human Language Technologies}, pages 180--189,
  Portland, Oregon, USA. Association for Computational Linguistics.

\bibitem[{Zhang et~al.(2019)Zhang, Sproat, Ng, Stahlberg, Peng, Gorman, and
  Roark}]{zhang-cl-textnorm}
Hao Zhang, Richard Sproat, Axel~H. Ng, Felix Stahlberg, Xiaochang Peng, Kyle
  Gorman, and Brian Roark. 2019.
\newblock \href {https://doi.org/10.1162/coli\_a\_00349} {Neural models of text
  normalization for speech applications}.
\newblock \emph{Computational Linguistics}, 45(2):293--337.

\bibitem[{Zhang and Lapata(2017)}]{zhang-lapata-2017-sentence}
Xingxing Zhang and Mirella Lapata. 2017.
\newblock \href {https://doi.org/10.18653/v1/D17-1062} {Sentence simplification
  with deep reinforcement learning}.
\newblock In \emph{Proceedings of the 2017 Conference on Empirical Methods in
  Natural Language Processing}, pages 584--594, Copenhagen, Denmark.
  Association for Computational Linguistics.

\bibitem[{Zhang and Weiss(2016)}]{zhang-weiss-2016-stack}
Yuan Zhang and David Weiss. 2016.
\newblock \href {https://doi.org/10.18653/v1/P16-1147} {Stack-propagation:
  Improved representation learning for syntax}.
\newblock In \emph{Proceedings of the 54th Annual Meeting of the Association
  for Computational Linguistics (Volume 1: Long Papers)}, pages 1557--1566,
  Berlin, Germany. Association for Computational Linguistics.

\bibitem[{Zhao et~al.(2019)Zhao, Wang, Shen, Jia, and
  Liu}]{zhao-etal-2019-improving}
Wei Zhao, Liang Wang, Kewei Shen, Ruoyu Jia, and Jingming Liu. 2019.
\newblock \href {https://doi.org/10.18653/v1/N19-1014} {Improving grammatical
  error correction via pre-training a copy-augmented architecture with
  unlabeled data}.
\newblock In \emph{Proceedings of the 2019 Conference of the North {A}merican
  Chapter of the Association for Computational Linguistics: Human Language
  Technologies, Volume 1 (Long and Short Papers)}, pages 156--165, Minneapolis,
  Minnesota. Association for Computational Linguistics.

\end{thebibliography}

\clearpage

\appendix

\section{Task-specific Tag Sets}
\label{sec:tags}

Tables \ref{tab:tag-vocab-textnorm} to \ref{tab:tag-vocab-errant} list the non-trivial tag sets for text normalization, sentence fusion, and grammatical error correction respectively. In addition to the tags listed in the tables, we use the tags \texttt{SELF} and \texttt{EOS} (end of sequence). For sentence splitting and simplification we use the trivial tag set consisting of \texttt{SELF}, \texttt{NON\_SELF}, and \texttt{EOS}.

\begin{table}[b!]
\centering
\small
\begin{tabularx}{\linewidth}{lX}
\hline \textbf{Tag} & \textbf{Description} \\ \hline
\texttt{PLAIN} &  Ordinary word \\
\texttt{PUNCT} &  Punctuation \\
\texttt{TRANS} &  Transliteration \\
\texttt{LETTERS} &  Letter sequence \\
\texttt{CARDINAL} &  Cardinal number \\
\texttt{VERBATIM} &  Verbatim reading of character sequence  \\
\texttt{ORDINAL} &  Ordinal number \\
\texttt{DECIMAL} &  Decimal fraction \\
\texttt{ELECTRONIC} &  Electronic address  \\
\texttt{DIGIT} &  Digit sequence \\
\texttt{MONEY} &  Currency amount \\
\texttt{FRACTION} &  Non-decimal fraction  \\
\texttt{TIME} &  Time expression \\
\texttt{ADDRESS} &  Street address \\
\hline
\end{tabularx}
\caption{\label{tab:tag-vocab-textnorm} Semiotic class tags for text normalization copied verbatim from the Table 3 caption of \citet{rws-rnn-text-norm}.}
\end{table}

\begin{table}[t!]
\centering
\small
\begin{tabularx}{\linewidth}{lX}
\hline \textbf{Tag} & \textbf{Description} \\ \hline
\texttt{PAIR\_ANAPHORA} & Anaphora \\
\texttt{PAIR\_CONN} & Discourse connective \\
\texttt{PAIR\_CONN\_ANAPHORA} & Discourse connective + anaphora \\
\texttt{PAIR\_NONE} & None (Control) \\
\texttt{SINGLE\_APPOSITION} & Apposition \\
\texttt{SINGLE\_CATAPHORA} & Cataphora \\
\texttt{SINGLE\_CONN\_INNER} & Inner connective \\
\texttt{SINGLE\_CONN\_INNER\_ANAPH.} & Inner connective + anaphora \\
\texttt{SINGLE\_CONN\_START} & Forward connective \\
\texttt{SINGLE\_RELATIVE} & Relative clause \\
\texttt{SINGLE\_S\_COORD} & Sentence coordination \\
\texttt{SINGLE\_S\_COORD\_ANAPHORA} & Sentence coordination + anaphora \\
\texttt{SINGLE\_VP\_COORD} & Verb phrase coordination \\
\hline
\end{tabularx}
\caption{\label{tab:tag-vocab-discofuse} DiscoFuse discourse types. The type descriptions are copied verbatim from Table 7 of \citet{geva-etal-2019-discofuse}. The \texttt{SINGLE} and \texttt{PAIR} prefixes indicate whether the input is a single sentence or two consecutive sentences.}
\end{table}

\begin{table}[t!]
\centering
\small
\begin{tabularx}{\linewidth}{lX}
\hline \textbf{Tag} & \textbf{Description} \\ \hline
\texttt{ADJ} & Adjective (``big'' $\rightarrow$ ``wide'') \\
\texttt{ADJ:FORM} & Comparative or superlative adjective errors. \\
\texttt{ADV} & Adverb (``speedily'' $\rightarrow$ ``quickly'') \\
\texttt{CONJ} & Conjunction (``and'' $\rightarrow$ ``but'') \\
\texttt{CONTR} & Contraction (``n't'' $\rightarrow$ ``not'') \\
\texttt{DET} & Determiner (``the'' $\rightarrow$ ``a'') \\
\texttt{MORPH} & Morphology \\
\texttt{NOUN} & Noun (``person'' $\rightarrow$ ``people'')  \\
\texttt{NOUN:INFL} & Noun inflection \\
\texttt{NOUN:NUM} & Noun number (``cat'' $\rightarrow$ ``cats'')  \\
\texttt{NOUN:POSS} & Noun possessive (``friends'' $\rightarrow$ ``friend's'')  \\
\texttt{ORTH} & Orthography case and/or whitespace errors. \\
\texttt{OTHER} & Other \\
\texttt{PART} & Particle (``(look) in'' $\rightarrow$ ``(look) at'')  \\
\texttt{PREP} & Preposition (``of'' $\rightarrow$ ``at'') \\
\texttt{PRON} & Pronoun (``ours'' $\rightarrow$ ``ourselves'')  \\
\texttt{PUNCT} & Punctuation (``!'' $\rightarrow$ ``.'')  \\
\texttt{SPELL} & Spelling (``genectic'' $\rightarrow$ ``genetic'', ``color'' $\rightarrow$ ``colour'')  \\
\texttt{UNK} & Unknown: The annotator detected an error but was unable to correct it.  \\
\texttt{VERB} & Verb (``ambulate'' $\rightarrow$ ``walk'')  \\
\texttt{VERB:FORM} & Verb form  \\
\texttt{VERB:INFL} & Verb inflection: misapplication of tense morphology.  \\
\texttt{VERB:SVA} & Subject-verb agreement (``(He) have'' $\rightarrow$ ``(He) has'')  \\
\texttt{VERB:TENSE} & Verb tense (includes inflectional and periphrastic tense, modal verbs and passivization).  \\
\texttt{WO} & Word order (``only can'' $\rightarrow$ ``can only'')  \\
\hline
\end{tabularx}
\caption{\label{tab:tag-vocab-errant} ERRANT tag vocabulary for grammatical error correction copied verbatim from Table 2 of \citet{bryant-etal-2017-automatic}.}
\end{table}

\section{Example Outputs}
\label{sec:examples}

Tables \ref{tab:examples-fusion} to \ref{tab:examples-gec} provide example outputs from the \emph{Seq2Edits} model. We use word-level rather than subword- or character-level source positions and collapse multi-word replacements into a single operation in the edit representation examples for clarity.

\begin{table*}
\centering
\small
\begin{tabularx}{\linewidth}{lX}
\hline
Source &  $_0$ Some $_1$ Haemulon $_2$ species $_3$ eat $_4$ plankton $_5$ in $_6$ the $_7$ open $_8$ water $_9$ . $_{10}$ Most $_{11}$ seek $_{12}$ small $_{13}$ prey $_{14}$ on $_{15}$ the $_{16}$ seabed $_{17}$ . $_{18}$ \\
Reference &  Some Haemulon species eat plankton in the open water , but most seek small prey on the seabed . \\ 
Edit model &  $\underbrace{\text{Some Haemulon species eat plankton in the open water}}_\mathtt{SELF}$ $\underbrace{\text{, but most}}_\mathtt{SINGLE\_S\_COORD}$ $\underbrace{\text{seek small prey on the seabed .}}_\mathtt{SELF}$ \\
Edits & (\texttt{SELF}, 9, \texttt{SELF}), (\texttt{SINGLE\_S\_COORD}, 11, `, but most'), (\texttt{SELF}, 18, \texttt{SELF}) \\
\hline
Source & $_0$ It $_1$ is $_2$ a $_3$ fan $_4$ favourite $_5$ . $_6$ It $_7$ has $_8$ been $_9$ played $_{10}$ in $_{11}$ almost $_{12}$ every $_{13}$ concert $_{14}$ to $_{15}$ date $_{16}$ since $_{17}$ its $_{18}$ initial $_{19}$ performance $_{20}$ . $_{21}$ \\
Reference &  Being a fan favourite , it has been played in almost every concert to date since its initial performance . \\ 
Edit model &  $\underbrace{\text{Being}}_\mathtt{SINGLE\_CATAPHORA}$ $\underbrace{\text{a fan favourite}}_\mathtt{SELF}$ $\underbrace{\text{, it}}_\mathtt{SINGLE\_CATAPHORA}$ $\underbrace{\text{has been played in almost every concert to date since its initial}}_\mathtt{SELF}$ $\underbrace{\text{performance .}}_\mathtt{SELF}$ \\
Edits & (\texttt{SINGLE\_CATAPHORA}, 2, `Being'), (\texttt{SELF}, 5, \texttt{SELF}), (\texttt{SINGLE\_CATAPHORA}, 7, `, it'), (\texttt{SELF}, 21, \texttt{SELF}) \\
\hline
Source &  $_0$ Boggs $_1$ was $_2$ decommissioned $_3$ on $_4$ 20 $_5$ March $_6$ 1946 $_7$ . $_8$ Boggs $_9$ was $_{10}$ sold $_{11}$ for $_{12}$ scrap $_{13}$ on $_{14}$ 27 $_{15}$ November $_{16}$ 1946 $_{17}$ . $_{18}$ \\
Reference & Boggs was decommissioned on 20 March 1946 and sold for scrap on 27 November 1946 . \\ 
Edit model &  $\underbrace{\text{Boggs was decommissioned on 20 March 1946}}_\mathtt{SELF}$ $\underbrace{\text{and}}_\mathtt{SINGLE\_VP\_COORD}$ $\underbrace{\text{sold for scrap on 27 November 1946}}_\mathtt{SELF}$ \\
Edits & (\texttt{SELF}, 7, \texttt{SELF}), (\texttt{SINGLE\_VP\_COORD}, 10, `and'), (\texttt{SELF}, 18, \texttt{SELF}) \\
\hline
Source & $_0$ The $_1$ river $_2$ Dulais $_3$ flows $_4$ through $_5$ the $_6$ village $_7$ . $_8$ The $_9$ river $_{10}$ Dulais $_{11}$ was $_{12}$ often $_{13}$ referred $_{14}$ to $_{15}$ as $_{16}$ the $_{17}$ Black $_{18}$ River $_{19}$ due $_{20}$ to $_{21}$ pollution $_{22}$ by $_{23}$ coal $_{24}$ dust $_{25}$ from $_{26}$ the $_{27}$ local $_{28}$ mining $_{29}$ industry $_{30}$ . $_{31}$ \\
Reference &  The river Dulais flows through the village and was often referred to as the Black River due to pollution by coal dust from the local mining industry . \\ 
Edit model &  $\underbrace{\text{The river Dulais}}_\mathtt{SELF}$ $\underbrace{\text{, which}}_\mathtt{SINGLE\_RELATIVE}$ $\underbrace{\text{was often referred to as the Black River due to pollution by coal dust from the}}_\mathtt{SELF}$ $\underbrace{\text{local mining industry}}_\mathtt{SELF}$ $\underbrace{\text{, flows through the village .}}_\mathtt{SINGLE\_RELATIVE}$ \\
Edits & (\texttt{SELF}, 3, \texttt{SELF}), (\texttt{SINGLE\_RELATIVE}, 11, `, which'), (\texttt{SELF}, 30, \texttt{SELF}), (\texttt{SINGLE\_RELATIVE}, 31, `flows through the village .') \\
\hline
Source &  $_0$ Baldwin $_1$ of $_2$ Bourcq $_3$ married $_4$ Morphia $_5$ and $_6$ Joscelin $_7$ of $_8$ Courtenay $_9$ married $_{10}$ a $_{11}$ daughter $_{12}$ of $_{13}$ Constantine $_{14}$ . $_{15}$ Morphia $_{16}$ is $_{17}$ a $_{18}$ daughter $_{19}$ of $_{20}$ Gabriel $_{21}$ of $_{22}$ Melitene $_{23}$ . $_{24}$ \\
Reference &  Baldwin of Bourcq married Morphia , a daughter of Gabriel of Melitene , and Joscelin of Courtenay married a daughter of Constantine . \\ 
Edit model &  $\underbrace{\text{Baldwin of Bourcq married Morphia}}_\mathtt{SELF}$ $\underbrace{\text{, a daughter of Gabriel of Melitene ,}}_\mathtt{SINGLE\_APPOSITION}$ $\underbrace{\text{and Joscelin of Courtenay married}}_\mathtt{SELF}$ $\underbrace{\text{a daughter of Constantine .}}_\mathtt{SELF}$  \\
Edits & (\texttt{SELF}, 5, \texttt{SELF}), (\texttt{SINGLE\_APPOSITION}, 5, `, a daughter of Gabriel of Melitene'), (\texttt{SELF}, 15, \texttt{SELF}), (\texttt{SINGLE\_APPOSITION}, 24, \texttt{DEL}) \\
\hline
\end{tabularx}
\caption{\label{tab:examples-fusion} Sentence fusion examples from the DiscoFuse dataset~\citep{geva-etal-2019-discofuse}.}
\end{table*}

\begin{table*}
\centering
\small
\begin{tabularx}{\linewidth}{lX}
\hline
Source &  $_0$ Jungle $_1$ Strike $_2$ " $_3$ Sega $_4$ Force $_5$ July $_6$ 93 $_7$ ( $_8$ issue $_9$ 19 $_{10}$ ) $_{11}$ , $_{12}$ pp $_{13}$ . $_{14}$  \\
Reference &  Jungle Strike sil Sega Force July ninety \textbf{three} sil issue nineteen sil sil p p sil \\ 
Full seq. &  Jungle Strike sil Sega Force july ninety \textbf{third} sil issue nineteen sil sil p p sil \\
Edit model &  $\underbrace{\text{Jungle Strike}}_\mathtt{SELF}$ $\underbrace{\text{sil}}_\mathtt{PUNCT}$ $\underbrace{\text{Sega Force July}}_\mathtt{SELF}$ $\underbrace{\text{ninety \textbf{three}}}_\mathtt{CARDINAL}$ $\underbrace{\text{sil}}_\mathtt{PUNCT}$ $\underbrace{\text{issue}}_\mathtt{SELF}$ $\underbrace{\text{nineteen}}_\mathtt{CARDINAL}$ $\underbrace{\text{sil}}_\mathtt{PUNCT}$ $\underbrace{\text{sil}}_\mathtt{PUNCT}$ $\underbrace{\text{p p}}_\mathtt{VERBATIM}$ $\underbrace{\text{sil}}_\mathtt{PUNCT}$   \\
Edits & (\texttt{SELF}, 2, \texttt{SELF}), (\texttt{PUNCT}, 3, `sil'), (\texttt{SELF}, 6, \texttt{SELF}), (\texttt{CARDINAL}, 7, `ninety three'), (\texttt{PUNCT}, 8, `sil'), (\texttt{SELF}, 9, \texttt{SELF}), (\texttt{CARDINAL}, 10, `nineteen'), (\texttt{PUNCT}, 11, `sil'), (\texttt{PUNCT}, 12, `sil'), (\texttt{VERBATIM}, 13, `p p'), (\texttt{PUNCT}, 14, `sil') \\
\hline
Source &  $_0$ Service $_1$ operates $_2$ from $_3$ the $_4$ Courthouse $_5$ at $_6$ 1030 $_7$ am $_8$ and $_9$ 1230 $_{10}$ pm $_{11}$ . $_{12}$\\
Reference &  Service operates from the Courthouse at \textbf{ten thirty} am and \textbf{twelve thirty} p m sil \\ 
Full seq. &  Service operates from the Courthouse at \textbf{one thousand thirty} a m and \textbf{one thousand two hundred thirty} p m sil \\
Edit model &  $\underbrace{\text{Service operates from the Courthouse at}}_\mathtt{SELF}$ $\underbrace{\text{\textbf{ten thirty}}}_\mathtt{DATE}$ $\underbrace{\text{am and}}_\mathtt{SELF}$ $\underbrace{\text{\textbf{twelve thirty}}}_\mathtt{DATE}$ $\underbrace{\text{p m}}_\mathtt{VERBATIM}$ $\underbrace{\text{sil}}_\mathtt{PUNCT}$ \\
Edits & (\texttt{SELF}, 6, \texttt{SELF}), (\texttt{DATE}, 7, `ten thirty'), (\texttt{SELF}, 9, \texttt{SELF}), (\texttt{DATE}, 10, `twelve thirty'), (\texttt{VERBATIM}, 11, `p m'), (\texttt{PUNCT}, 12, `sil') \\
\hline
Source &  $_0$ 168 $_1$ ( $_2$ November $_3$ 1991 $_4$ ) $_5$ , $_6$ pp $_7$ . $_8$ \\
Reference &  \textbf{one hundred sixty eight} sil november nineteen ninety one sil sil p p sil \\ 
Full seq. &  \textbf{thousand one hundred sixty eight} sil november nineteen ninety one sil sil p p sil \\
Edit model&  $\underbrace{\text{\textbf{one hundred sixty eight}}}_\mathtt{CARDINAL}$ $\underbrace{\text{sil}}_\mathtt{PUNCT}$ $\underbrace{\text{november nineteen ninety one}}_\mathtt{DATE}$ $\underbrace{\text{sil}}_\mathtt{PUNCT}$ $\underbrace{\text{sil}}_\mathtt{PUNCT}$ $\underbrace{\text{p p}}_\mathtt{VERBATIM}$ $\underbrace{\text{sil}}_\mathtt{PUNCT}$ \\
Edits & (\texttt{CARDINAL}, 1, `one hundred sixty eight'), (\texttt{PUNCT}, 2, `sil'), (\texttt{DATE}, 4, `november nineteen ninety one'), (\texttt{PUNCT}, 5, `sil'), (\texttt{PUNCT}, 6, `sil'), (\texttt{VERBATIM}, 7, `p p'), (\texttt{PUNCT}, 8, `sil') \\
\hline
\end{tabularx}
\caption{\label{tab:examples-textnorm} English text normalization examples from the dataset provided by \citet{rws-rnn-text-norm}.}
\end{table*}

Example outputs of our sentence fusion system are shown in Table~\ref{tab:examples-fusion}. The predicted tags capture the variety of strategies for sentence fusion, such as simple connector particles (\texttt{SINGLE\_S\_COORD}), cataphoras (\texttt{SINGLE\_CATAPHORA}), verb phrases (\texttt{SINGLE\_VP\_COORD}), relative clauses (\texttt{SINGLE\_RELATIVE}), and appositions (\texttt{SINGLE\_APPOSITION}). The last example in Table~\ref{tab:examples-fusion} demonstrates that our model is able to produce even major rewrites.

\begin{table*}[t!]
\centering
\small
\begin{tabularx}{\linewidth}{lX}
\hline
Source &  $_0$ It $_1$ will $_2$ be $_3$ very $_4$ cool $_5$ to $_6$ see $_7$ the $_8$ \textbf{las} $_9$ \textbf{part} $_{10}$ \textbf{mokingjay} $_{11}$ ! $_{12}$ \\
Reference &  It will be very cool to see the \textbf{last part of Mokingjay} ! \\ 
Full seq. &  It will be very cool to see the \textbf{last mokingjay} ! \\
Edit model &  $\underbrace{\text{It will be very cool to see the}}_\mathtt{SELF}$ $\underbrace{\text{\textbf{last}}}_\mathtt{SPELL}$ $\underbrace{\text{\textbf{part}}}_\mathtt{SELF}$ $\underbrace{\text{\textbf{of}}}_\mathtt{PREP}$ $\underbrace{\text{\textbf{mokingjay} !}}_\mathtt{SELF}$ \\
Edits & (\texttt{SELF}, 8, \texttt{SELF}), (\texttt{SPELL}, 9, `last'), (\texttt{SELF}, 10, \texttt{SELF}), (\texttt{PART}, 10, `of'), (\texttt{SELF}, 12, \texttt{SELF}) \\
\hline
Source &  $_0$ If $_1$ she $_2$ was $_3$ n't $_4$ awake $_5$ , $_6$ why $_7$ \textbf{she} $_8$ \textbf{could} $_9$ \textbf{n't} $_{10}$ remember $_{11}$ anything $_{12}$ after $_{13}$ that $_{14}$ ? $_{15}$ \\
Reference &  If she was n't awake , why \textbf{could n't} she remember anything after that ? \\ 
Full seq. &  If she was n't awake , why \textbf{she could n't} remember anything after that ?  \\
Edit model &  $\underbrace{\text{If she was n't awake , why}}_\mathtt{SELF}$ $\underbrace{\text{\textbf{could n't she}}}_\mathtt{WO}$ $\underbrace{\text{remember anything after that ?}}_\mathtt{SELF}$ \\
Edits & (\texttt{SELF}, 7, \texttt{SELF}), (\texttt{WO}, 10, `could n't she'), (\texttt{SELF}, 15, \texttt{SELF}) \\
\hline
Source & $_0$ \textbf{Less} $_1$ channels $_2$ means $_3$ \textbf{less} $_4$ choices $_5$ . $_6$ \\
Reference &  \textbf{Fewer} channels means \textbf{fewer} choices . \\ 
Full seq. &  \textbf{Less} channels means \textbf{fewer} choices .  \\
Edit model &   $\underbrace{\text{\textbf{Fewer}}}_\mathtt{ADJ}$ $\underbrace{\text{channels means}}_\mathtt{SELF}$ $\underbrace{\text{\textbf{fewer}}}_\mathtt{ADJ}$ $\underbrace{\text{choices .}}_\mathtt{SELF}$ \\
Edits & (\texttt{ADJ}, 1, `Fewer'), (\texttt{SELF}, 3, \texttt{SELF}), (\texttt{ADJ}, 4, `fewer'), (\texttt{SELF}, 6, \texttt{SELF}) \\
\hline
Source & $_0$ On $_1$ the $_2$ one $_3$ hand $_4$ travel $_5$ by $_6$ car $_7$ are $_8$ really $_9$ much $_{10}$ more $_{11}$ convenient $_{12}$ as $_{13}$ \textbf{give} $_{14}$ the $_{15}$ chance $_{16}$ \textbf{to} $_{17}$ \textbf{you} $_{18}$ to $_{19}$ be $_{20}$ independent $_{21}$ . $_{22}$ \\
Reference &  On the one hand \textbf{,} travel by car \textbf{is} really much more convenient \textbf{,} as \textbf{it gives you} the chance to be independent . \\ 
Full seq. &  On the one hand \textbf{,} travel by car \textbf{is} really much more convenient \textbf{,} as \textbf{it gives you} the chance to be independent .  \\
Edit model &  $\underbrace{\text{On the one hand}}_\mathtt{SELF}$ $\underbrace{\text{\textbf{,}}}_\mathtt{PUNCT}$ $\underbrace{\text{travel by car}}_\mathtt{SELF}$ $\underbrace{\text{\textbf{is}}}_\mathtt{VERB:SVA}$ $\underbrace{\text{really much more convenient}}_\mathtt{SELF}$ $\underbrace{\text{\textbf{,}}}_\mathtt{PUNCT}$ $\underbrace{\text{\textbf{as give}}}_\mathtt{SELF}$ $\underbrace{\text{\textbf{you}}}_\mathtt{PRON}$ $\underbrace{\text{the chance}}_\mathtt{SELF}$ $\underbrace{\text{to be independent .}}_\mathtt{SELF}$  \\
Edits & (\texttt{SELF}, 4, \texttt{SELF}), (\texttt{PUNCT}, 4, `,'), (\texttt{SELF}, 7, \texttt{SELF}), (\texttt{VERB:SVA}, 8, `is'), (\texttt{SELF}, 12, \texttt{SELF}), (\texttt{PUNCT}, 12, `,'), (\texttt{SELF}, 14, \texttt{SELF}), (\texttt{PRON}, 14, `you'), (\texttt{SELF}, 16, \texttt{SELF}), (\texttt{PRON}, 18, \texttt{DEL}), (\texttt{SELF}, 22, \texttt{SELF}) \\
\hline
\end{tabularx}
\caption{\label{tab:examples-gec} Grammatical error correction examples from BEA-dev~\citep{bryant-etal-2019-bea}.}
\end{table*}

Table \ref{tab:examples-textnorm} compares our model with a full sequence baseline on English text normalization. Correctly predicting non-trivial tags helps our model to choose the right verbalizations. In the first example in Table \ref{tab:examples-textnorm}, our model predicts the \texttt{CARDINAL} tag rather than \texttt{ORDINAL} and thus produces the correct verbalization for `93'. In the second example, our model generates a time expression for `1030' and `1230' as it predicted the \texttt{DATE} tag for these spans. The third example demonstrates that the edit model can avoid some of the `unrecoverable' errors~\citep{rws-rnn-text-norm} of the full sequence model such as mapping `168' to `thousand one hundred sixty eight'.

Finally, the grammatical error correction examples in Table~\ref{tab:examples-gec} demonstrate the practical advantage of predicting tags along with the edits as they provide useful feedback to the user. The second example in Table~\ref{tab:examples-gec} shows that our model is able to handle more complex operations such as word reorderings. However, our model fails to inflect ``give'' correctly in the last example, suggesting that one weakness of our edit model compared to a full sequence model is a weaker target side language model resulting in less fluent output. This issue can be mitigated by using stronger models e.g.\ this particular issue is fixed in our ensemble.

\end{document}